\documentclass{article}

\newif\ifdraft
\draftfalse

\usepackage[table]{xcolor}
\usepackage[normalem]{ulem}

\usepackage{PRIMEarxiv}

\usepackage[utf8]{inputenc} 
\usepackage[T1]{fontenc}    
\usepackage{hyperref}       
\usepackage{url}            
\usepackage{booktabs}       
\usepackage{amsfonts}       
\usepackage{nicefrac}       
\usepackage{microtype}      
\usepackage{lipsum}
\usepackage{fancyhdr}       
\usepackage{graphicx}       
\graphicspath{{media/}}     
\usepackage{multirow}
\usepackage{amsmath} 
\usepackage{comment}
\usepackage{caption}

\usepackage{booktabs}
\usepackage{multirow}
\usepackage{tabularx}
\usepackage{makecell}

\usepackage[normalem]{ulem} 
\newcommand{\stkout}[1]{\ifmmode\text{\sout{\ensuremath{#1}}}\else\sout{#1}\fi}

\ifdraft

\newcommand{\added}[1]{\textcolor{blue}{#1}}
\newcommand{\deleted}[1]{\textcolor{red}{\stkout{#1}}}
\newcommand{\replaced}[2]{\textcolor{blue}{#1} \textcolor{red}{\stkout{#2}}}
\newcommand{\deletedfloat}[1]{}
\newcommand{\commented}[1]{\textcolor{blue}{#1}}
\else

\newcommand{\added}[1]{#1}
\newcommand{\deleted}[1]{}
\newcommand{\replaced}[2]{#1}
\newcommand{\deletedfloat}[1]{}
\newcommand{\commented}[1]{}
\fi

\title{Uncertainty reliability under domain shift: an investigation for data-driven blood pressure estimation in photoplethysmography
}

\author{
Mohammad Moulaeifard$^{\ast}$,
Ciaran Bench$^{\dagger}$,
Philip J.~Aston$^{\dagger\ddagger}$,
Nils~Strodthoff$^{\ast\diamond}$ \\
\\
$^{\ast}$AI4Health Department, University of Oldenburg, Oldenburg, Germany \\
$^{\dagger}$Department of Data Science and AI, National Physical Laboratory, Teddington, United Kingdom \\
$^{\ddagger}$School of Mathematics and Physics, University of Surrey, Surrey, United Kingdom \\
\\
$^{\diamond}$\textit{Corresponding author:} \texttt{nils.strodthoff@uol.de}
}

\begin{document}
\maketitle

\begin{abstract}
Uncertainty quantification (UQ) is critical for safety-critical domains like healthcare, yet it is rarely evaluated under realistic out-of-distribution (OOD) conditions. Here, we assessed predictive performance and uncertainty reliability for deep learning-based blood pressure (BP) estimation from photoplethysmography (PPG) signals under both in-distribution (ID) and OOD settings. Using an XResNet1D-50 trained on PulseDB and tested on four external datasets, we compared deep ensembles (DE) and Monte Carlo dropout (MCD) with Gaussian negative log-likelihood (GNLL) and mean squared error (MSE) losses, optionally followed by post-hoc recalibration via conformal prediction (CP), temperature scaling (TS), and isotonic regression (IR).

The key findings of our study are as follows: (1) DE provides stronger predictive robustness under domain shift than MCD, an advantage that becomes clear primarily under external shift. (2) Recalibrated GNLL-based methods yield the best uncertainty calibration (e.g., GNLL+DE+CP for systolic blood pressure (SBP), GNLL+DE+TS for diastolic blood pressure (DBP)), while MSE-based uncertainty requires recalibration to become practically useful. (3) Across settings, CP and TS offer the most consistent gains, with IR remaining competitive in several cases.

Overall, our results identify DE-based methods as most robust for predictive performance under domain shift, GNLL as strongest for native UQ, and recalibration as essential for making MSE-based uncertainty practical. These findings highlight the need to jointly assess predictive accuracy and calibration on external data for trustworthy cuffless BP estimation.

\end{abstract}


\section{Introduction}

\paragraph{Wearable PPG for BP monitoring}
Continuous BP monitoring using wearable sensors is crucial, as cuff-based BP measurement is not suited for continuous BP monitoring during daily activities, stress, and sleep \cite{mukkamala2015toward, elgendi2019cuffless}.Given the large amounts of physiological data generated during continuous monitoring, deep learning (DL) approaches have enabled end-to-end modeling of raw PPG signal time-series, outperforming traditional feature-based approaches that rely on hand-crafted pulse morphology descriptors \cite{slapnicar2019blood, wang2018deep}. In addition, the existing benchmarking studies have shown that convolutional neural networks operating on raw PPG signals consistently achieve state-of-the-art performance for BP estimation across multiple datasets and input representations \cite{moulaeifard2025mlppg,koparir2024cuffless}. However, these results are typically obtained under controlled or dataset-specific evaluation settings, and do not necessarily reflect model behaviour when deployed on previously unseen data or under real-world measurement variability. \emph{This naturally raises the question: to what extent can we rely on these performance results?}

\paragraph{UQ for reliable BP estimation}
In the context of healthcare applications, accuracy alone is insufficient for ensuring the safe application of the model, as performance reported on a test set does not provide
information about the confidence or reliability of individual predictions. As a result, predictive accuracy by itself provides limited insight into the trustworthiness of individual predictions when models are deployed in real-world settings. UQ provides valuable insights into model confidence of the predictions and enables risk-informed decision-making \cite{kendall2017uncertainties,kompa2021confidence}. For PPG-based BP estimation, uncertainty estimation is particularly crucial due to motion artifacts, sensor variability, and physiological heterogeneity encountered in real-world settings that can have a negative impact on performance \cite{elgendi2019cuffless,abdar2021review}. 

In the context of deep learning models, several sources of uncertainty may be present. These are commonly categorised into aleatoric uncertainty, which arises from inherent and irreducible variability in the data, and epistemic uncertainty, which reflects uncertainty in the model and can be reduced with additional data or improved modelling \cite{kendall2017uncertainties}. For regression tasks such as BP estimation, aleatoric uncertainty is typically modelled using probabilistic regression objectives. There are a large number of methods used to model epistemic uncertainty, such as MCD \cite{gal2016dropout}, and DE \cite{lakshminarayanan2017simple}.

Recent works have investigated uncertainty-aware modelling for wearable PPG tasks. For instance, \cite{bench2025uncertainty} explored approximate variational inference and dropout-based approaches for UQ in wearable PPG prediction tasks, including BP regression, 
Also, \cite{bench2026systematic} conducted a subsequent systematic evaluation comparing multiple UQ techniques for PPG signal analysis. 

\paragraph{Generalizability under domain shift}\label{par:ood_generalization}
In real-world scenarios, wearable PPG models are frequently applied to data that differ from the training distribution due to changes in sensors, acquisition protocols, or target populations. Prior benchmarking studies have shown the lack of robustness under such OOD conditions, e.g., \cite{moulaeifard2025benchmark}. 
Recent studies outside of PPG analysis, e.g.\ \cite{d2022underspecification, ramos2025underspecification}, have highlighted the challenge of underspecification in modern machine learning, and it's impact on model generalizability. This term describes the situation in which multiple predictors derived from the same model and training pipeline achieve nearly identical ID performance, while showing markedly different behaviour on OOD data. To the best of our knowledge, this issue has not yet been explored in PPG-based prediction tasks.



\paragraph{UQ for PPG-based BP estimation}
UQ aims to estimate the degree of doubt in a model's prediction, and may provide an indication of its trustworthiness without prior knowledge of the ground truth. Aleatoric and epistemic uncertainty are the two most significant sources of uncertainty is most practical measurement scenarios. Aleatoric uncertainty (irreducible data uncertainty) is often modelled using likelihood-based losses (e.g., the Gaussian negative log likelihood). Epistemic uncertainty (reducible model uncertainty) may be approximated through stochastic inference techniques such as MCD \cite{gal2016dropout} or DE \cite{lakshminarayanan2017simple}.

Several studies have incorporated uncertainty-aware modelling into PPG-based prediction tasks, including BP estimation \cite{liu2022videocad,asgharnezhad2023improving,han2023noncontact,chen2022quantifying,das2022bayesbeat}. This includes models trained on data acquired from wearable devices, e.g., \cite{bench2025uncertainty} explored approximate variational inference approaches to UQ. They demonstrated that the reliability of predicted uncertainties is highly sensitive to modelling choices, and the evaluation metrics considered. In a subsequent systematic study, the same authors compared a wide range of UQ techniques for modelling uncertainties in PPG signal analyses, and their results show varied levels of calibration performances with respect to uncertainties and errors \cite{bench2026systematic}. However, these PPG-focused studies have primarily evaluated uncertainty under ID conditions, leaving open questions about reliability under domain shift.

\paragraph{Uncertainty reliability under dataset shift}
Existing works considering non-PPG related prediction tasks have shown that uncertainties estimated from OOD data (even from internal splits of a given dataset) may be unreliable \cite{ovadia2019can,nado2021evaluating}. These issues were observed across all types of techniques for estimating predictive uncertainty, including Bayesian neural networks, DE, MCD, and post-hoc calibration methods. In particular, post-hoc recalibration methods have been shown to exhibit poorer generalisability under distribution
shift \cite{ovadia2019can}), increasing concerns associated with reliability in instances under distributional mismatch, particularly for regression \cite{gawlikowski2021survey}.

Our study investigates the reliability and generalizability of uncertainty estimates for data-driven BP estimation from PPG signals. By employing the established ID and OOD benchmarks from \cite{moulaeifard2025benchmark}, previously evaluated without UQ, this study seeks to assess whether widely used UQ methods provide reliable uncertainty estimates when models are exposed to domain shift. 

\paragraph{Technical contributions}
The technical contributions of this study are as follows:
\begin{itemize}
\item A comprehensive benchmark of predictive accuracy and uncertainty reliability for PPG-based BP estimation under both ID and OOD conditions.
\item A systematic comparison of uncertainty estimation strategies (MCD vs.\ DE), probabilistic losses (GNLL vs.\ MSE) , and post-hoc recalibration methods such as CP, TS, and IR for uncertainty estimation in cuffless BP prediction.
\item A detailed analysis of how uncertainty reliability degrades under distributional shift, and of how probabilistic modeling, ensembling, and recalibration affect robustness across external datasets.
\end{itemize}

\section{Materials \& methods}

\subsection{Datasets}
\label{sec:datasets}

\paragraph{ID Dataset} We built on the work described in \cite{moulaeifard2025benchmark}, which used the PulseDB dataset \cite{wang2023pulsedb} for training, as well as a collection of external datasets introduced in \cite{gonzalez2023benchmark} for OOD evaluation. PulseDB includes selected, pre-processed PPG waveforms derived from the MIMIC-III \cite{moody2020mimic} and VitalDB \cite{lee2022vitaldb} databases. \added{It is worth noting that \cite{moulaeifard2025benchmark} identified VitalDB-based training subsets, including CalibFree-VitalDB, i.e., VitalDB-derived scenario with no shared patients between training and test data, as important contributors to improved OOD generalization. Motivated by these results, we used CalibFree-VitalDB as the (ID) training dataset in our work.}

\paragraph{OOD Datasets} The OOD datasets introduced in \cite{gonzalez2023benchmark} are diverse in nature and acquisition settings, including differences in recording environment, subject population, and signal duration. These data include Sensors \added{\cite{aguirre2021blood}}, derived from intensive care unit (ICU) recordings in the MIMIC-III database; CI \added{(alternatively known as the Cuff-Less Blood Pressure Estimation dataset) \cite{kachuee2015cuff}}, extracted from the MIMIC-II database, which is the largest among the external datasets. On the other hand, the BCG \added{(bed-based ballistocardiography) \cite{carlson2020bed}} dataset was gathered in a controlled laboratory environment from a limited number of mostly healthy participants, whereas the PPGBP dataset \added{\cite{liang2018new}} consists of short PPG recordings collected from individuals with documented cardiovascular diseases. We briefly summarized the datasets utilized in this study in Table ~\ref{tab:dataset}. Detailed descriptions of the pre-processing procedures and dataset characteristics are provided in \cite{gonzalez2023benchmark}.

\begin{table*}[h]
\centering
\caption{Summary of the datasets utilized in this study: The CalibFree-VitalDB subset of PulseDB is used for training, and four external datasets are used for OOD evaluation.}
\scalebox{0.85}{
\begin{tabular}{lcccccc}

& \multicolumn{1}{c}{\textbf{ID dataset}} & \multicolumn{4}{c}{\textbf{OOD dataset}} \\ 
\cmidrule(lr){2-2} \cmidrule(lr){3-6}
\textbf{Metric} & \textbf{CalibFree-VitalDB} & \textbf{Sensors} & \textbf{UCI} & \textbf{BCG} & \textbf{PPGBP} \\
\midrule
\textbf{Unique subjects} & 1,158 & 1,195 & 10,793 & 40 & 218 \\
\textbf{Total Duration (h)} & $\sim$1,453 & $\sim$15 & $\sim$570 & $\sim$4 & $<$1 \\
\textbf{Segments (count , length)} & 523,080 , 10s & 11,102 , 5s & 410,596 , 5s & 3,063 , 5s & 619 , 2.1s \\
\textbf{Age (years, mean ± SD)} & 58.89 ± 15.07 & 57.1 ± 14.2 & unknown & 34.2 ± 14.5 & 56.9 ± 15.8 \\
\textbf{Gender (\%  Female)} & 42.08 & 40.2 & unknown & 55.5 & 53.1 \\
\textbf{SBP (mmHg, mean ± SD)} & 115.47 ± 18.91 & 134.36 ± 21.78 & 131.57 ± 11.16 & 120.99 ± 15.29 & 128.02 ± 20.50 \\
\textbf{DBP (mmHg, mean ± SD)} & 62.93 ± 12.06 & 65.37 ± 10.51 & 66.79 ± 10.48 & 67.23 ± 9.30 & 71.91 ± 11.20 \\
\bottomrule
\end{tabular}
}
\label{tab:dataset}
\end{table*}

It is worth noting that the ID training dataset is further partitioned into four disjoint splits: training (416{,}880 segments from 1{,}158 subjects), validation (32{,}400 segments from 90 subjects), calibration (16{,}200 segments from 45 subjects), and test (57{,}600 segments from 144 subjects). All examples in the external test sets were used for testing.

\subsection{UQ Techniques}
\label{sec:uq_methods}

\added{\added{Let $\mathbf{x}$ denote an input PPG segment and $y$ the corresponding ground truth BP value. In the MSE-based models, the network outputs a single scalar prediction $\hat{y}$, which can be interpreted as the predicted mean $\mu(\mathbf{x})$. In contrast, in the GNLL-based models, the network is parameterized to output two quantities, namely the predicted mean $\mu(\mathbf{x})$ and the predicted variance $\sigma^2(\mathbf{x})$, which together define a Gaussian predictive distribution. In this work, UQ was addressed through three complementary components:}}
\begin{enumerate}
    \item \textbf{Aleatoric uncertainty.} We modeled aleatoric uncertainty through the training objective using GNLL \cite{kendall2017uncertainties}, which enables direct prediction of both the target mean and variance. Under this formulation, the network predicts $\mu(\mathbf{x})$ and $\sigma^2(\mathbf{x})$, where $\mu(\mathbf{x})$ represents the predicted BP value and $\sigma^2(\mathbf{x})$ is interpreted as an input-dependent estimate of predictive uncertainty. The GNLL loss is given by
    \begin{equation}
    \mathcal{L}_{\mathrm{GNLL}} =
    \frac{1}{2}\log \sigma^2(\mathbf{x}) +
    \frac{(y - \mu(\mathbf{x}))^2}{2\sigma^2(\mathbf{x})}.
    \end{equation}
    
    This formulation allows the model to represent heteroscedastic uncertainty, such that samples associated with greater inherent ambiguity may be assigned larger predictive variances. \added{Models were also trained using the standard MSE loss, which remains widely used in PPG-based BP estimation \cite{wang2018deep}. In this case, the model is parameterized to output only a single point prediction $\hat{y}=\mu(\mathbf{x})$, and no input-dependent variance is estimated}
    
    \begin{equation}
    \mathcal{L}_{\mathrm{MSE}} = (y - \hat{y})^2 = (y - \mu(\mathbf{x}))^2.
    \end{equation}

    \item \textbf{Epistemic uncertainty.} In this study, epistemic uncertainty approximated using two model-based approaches, namely Monte Carlo Dropout (MCD) and Deep Ensembles (DE).

    \paragraph{Monte Carlo Dropout.} In this approach, dropout remains active at inference time, and each input sample is passed through the network $T$ times to obtain a set of stochastic predictions $\{\hat{y}^{(t)}\}_{t=1}^{T}$ \cite{gal2016dropout}. The variability across these forward passes is used as an estimate of predictive uncertainty. Specifically, the predictive variance is computed as
    \begin{equation}
    \mathrm{Var}_{\mathrm{MCD}}(\hat{y}) =
    \frac{1}{T}\sum_{t=1}^{T}
    \left(
    \hat{y}^{(t)} -
    \frac{1}{T}\sum_{t=1}^{T}\hat{y}^{(t)}
    \right)^2.
    \end{equation}
    We evaluated dropout rates of 0\%, 4\%, and 40\% in order to examine how the degree of stochasticity affects uncertainty estimation. Also, to mitigate variability arising from underspecification, each configuration was trained using five different random seeds. Each seed was treated as an independent model instance, and the resulting performance was summarized using the median and interquartile range (IQR) across seeds.
    
    \paragraph{Deep Ensembles.}
    Epistemic uncertainty was also estimated using DE composed of $K=5$ independently trained models with different random initializations \cite{lakshminarayanan2017simple}. For a given input $\mathbf{x}$, each ensemble member $k$ predicts a mean $\mu_k(\mathbf{x})$ and, in the GNLL setting, a variance $\sigma_k^2(\mathbf{x})$. The ensemble predictive mean was computed as
    \begin{equation}
    \mu_{\mathrm{ens}}(\mathbf{x}) =
    \frac{1}{K}\sum_{k=1}^{K}\mu_k(\mathbf{x}),
    \end{equation}
    and the epistemic uncertainty was estimated from the dispersion across ensemble predictions:
    \begin{equation}
    \sigma^2_{\mathrm{epi}}(\mathbf{x}) =
    \frac{1}{K}\sum_{k=1}^{K}
    \left(
    \mu_k(\mathbf{x}) - \mu_{\mathrm{ens}}(\mathbf{x})
    \right)^2.
    \end{equation}
    For GNLL-based models, the aleatoric uncertainty was computed as the average predicted variance across ensemble members:
    \begin{equation}
    \sigma^2_{\mathrm{ale}}(\mathbf{x}) =
    \frac{1}{K}\sum_{k=1}^{K}\sigma_k^2(\mathbf{x}).
    \end{equation}

    In both MCD and DE settings, the final predictive variance was obtained by combining aleatoric and epistemic components, such that
    \begin{equation}
    \sigma^2_{\mathrm{total}}(\mathbf{x}) = \sigma^2_{\mathrm{ale}}(\mathbf{x}) + \sigma^2_{\mathrm{epi}}(\mathbf{x}).
    \end{equation}
    
    \item \textbf{Post-hoc recalibration.} Post-hoc recalibration was applied after training to improve the calibration of the resulting uncertainty estimates. TS \cite{guo2017calibration}, IR \cite{robertson1988order}, and CP \cite{angelopoulos2023conformal} operated after model training and use a separate calibration set to improve the reliability of uncertainty estimates without altering the underlying network parameters.
    TS performs a global parametric rescaling of the predicted variance, while IR learns a monotonic mapping from predicted variance to empirical squared residuals. Both methods operate directly on variance estimates produced by GNLL or model-based approaches (MCD/DE). In contrast, CP does not recalibrate variance explicitly. Instead, it constructs prediction intervals based on normalized residuals computed on a calibration set, yielding intervals with finite-sample coverage guarantees under the assumption of exchangeability \cite{vovk2005algorithmic}. As a result, CP is naturally assessed using interval-based criteria, whereas TS and IR primarily act on continuous uncertainty estimates derived from probabilistic or ensemble-based predictions.
\end{enumerate}

\subsection{Training Protocol and Evaluation Metrics}
\label{sec:training_protocol}

\paragraph{Training protocol.}
Across all experiments, an effective batch size of 512 was achieved through gradient accumulation. All networks were trained for 50 epochs using the AdamW optimizer with a fixed learning rate of $10^{-3}$. To reduce overfitting, model selection was based on validation performance: the validation MAE was monitored throughout training, and the checkpoint achieving the lowest validation MAE within the 50 training epochs was retained for subsequent evaluation.

The models were trained using the native temporal resolution of each dataset, corresponding to 1{,}250 time steps (10\,s) for PulseDB, 625 time steps (5\,s) for the BCG, UCI, and Sensors datasets, and 262 time steps (2.1\,s) for PPGBP. Because all architectures employed global average pooling, the trained networks were able to accommodate variable-length input sequences during inference.

\paragraph{Accuracy evaluation metric.}
Consistent with standard practice in BP estimation, predictive accuracy was primarily quantified using the mean absolute error (MAE), defined as
\begin{equation}
\mathrm{MAE} = \frac{1}{n} \sum_{i=1}^{n} \left| \hat{y}_i - y_i \right|,
\end{equation}
where $n$ denotes the total number of predictions, $\hat{y}_i$ is the predicted BP value, and $y_i$ is the corresponding reference measurement.

\paragraph{Uncertainty evaluation metric.}
The quality of uncertainty estimates was evaluated using the Winkler score \cite{gneiting2007strictly}, which jointly reflects calibration and sharpness by rewarding narrow prediction intervals while penalizing intervals that fail to contain the true target value. For a central $(1-\alpha)$ prediction interval with lower bound $L$, upper bound $U$, and true target $y$, the Winkler score is defined as
\begin{equation}
W_{\alpha}(L,U;y)=
\begin{cases}
(U-L) + \dfrac{2}{\alpha}(L-y), & y < L, \\[6pt]
(U-L), & L \le y \le U, \\[6pt]
(U-L) + \dfrac{2}{\alpha}(y-U), & y > U.
\end{cases}
\end{equation}
Lower Winkler scores indicate better uncertainty estimates, reflecting a more favourable trade-off between interval sharpness and empirical coverage. Prediction intervals were evaluated at both the 1$\sigma$ and 2$\sigma$ levels, corresponding to nominal coverage levels of approximately 68\% and 95\%, respectively, under a Gaussian assumption. The final calibration summary score was defined as an equal-weight combination of the interval scores at these two levels, thereby balancing sharpness and coverage across different uncertainty regimes.

\paragraph{Bootstrapped confidence intervals.} We assessed the statistical uncertainty due to the finiteness and particular composition of the test set through \added{segment-level} bootstrapping on the test set \added{(consistent with our segment-level MAE and calibration evaluations)}. For each model and dataset (ID, BCG, Sensors, UCI, PPGBP), 1{,}000 bootstrap resamples were generated by sampling the test set with replacement, \added{with each resample containing the same number of segments as the original test set}. MAE and the calibration summary score (average of interval scores at 1$\sigma$ and 2$\sigma$) were recomputed for each resample, and results were summarized using the mean and 95\% confidence intervals.

For statistical comparison, we bootstrapped the performance difference between two models under consideration and considered the performance difference as statistically significant if the 95\% confidence interval for the performance difference excludes zero. We then designate the best-performing model as well as all models that are not statistically worse than it as \textbf{Tier~1}, all remaining models as \textbf{Tier~2}. For compactness, only the three best-performing Tier~2 methods are reported.

\subsection{Prediction Model Architecture}
\label{sec:uq_model}

In this work, we employed an MCD-enabled XResNet1D architecture, denoted as \texttt{xresnet1d50\_MCD}, to perform uncertainty-aware prediction \cite{gal2016dropout}.

\paragraph{Backbone: XResNet1D-50}
The base architecture follows the XResNet1D \cite{strodthoff2020deep} design, a one-dimensional adaptation of the ResNet architecture with improved initialization and operation ordering \cite{he2016resnet,howard2020fastai}. The network consists of a three-layer convolutional stem with strided convolutions followed by max pooling, four residual stages with block configuration $[3, 4, 6, 3]$, bottleneck residual blocks with an expansion factor of $4$ and a global pooling layer, and a fully connected prediction head. This design enables deep temporal feature extraction from one-dimensional physiological signals while maintaining stable optimization.

The structure of the residual block follows the conventional bottleneck design, consisting of a $1 \times 1$ convolution for dimension reduction, a $3 \times 3$ convolution for extracting temporal features, followed by a final $1 \times 1$ convolution for dimension enhancement. Batch normalization followed by ReLU activation is applied following the conventional design for ResNets, described in detail in \cite{he2016resnet}; BatchZero initialization is applied for stability improvement in the final convolution layers of each block.

\paragraph{MCD Integration}
Epistemic uncertainty estimation is enabled by incorporating dropout into the residual blocks in a principled and controlled manner. Inspired by prior analyses on dropout placement in residual networks, which highlight the importance of applying dropout at the block level to preserve optimization stability \cite{kim2023dropout}, dropout is applied only once per residual block, positioned after the residual convolutional mapping and before the residual addition, while no dropout is introduced within individual convolutional layers. Formally, each residual block computes:

\begin{equation}
    \mathbf{y} = \mathrm{ReLU}\big(\mathrm{Dropout}(F(\mathbf{x})) + \mathbf{x}\big),
\end{equation}
where $F(\cdot)$ denotes the residual mapping.

\section{Results}

\subsection{Quantifying distribution shifts}
\label{sec:res_ood}

\begin{figure}[h]
    \centering
    \includegraphics[width=\linewidth]{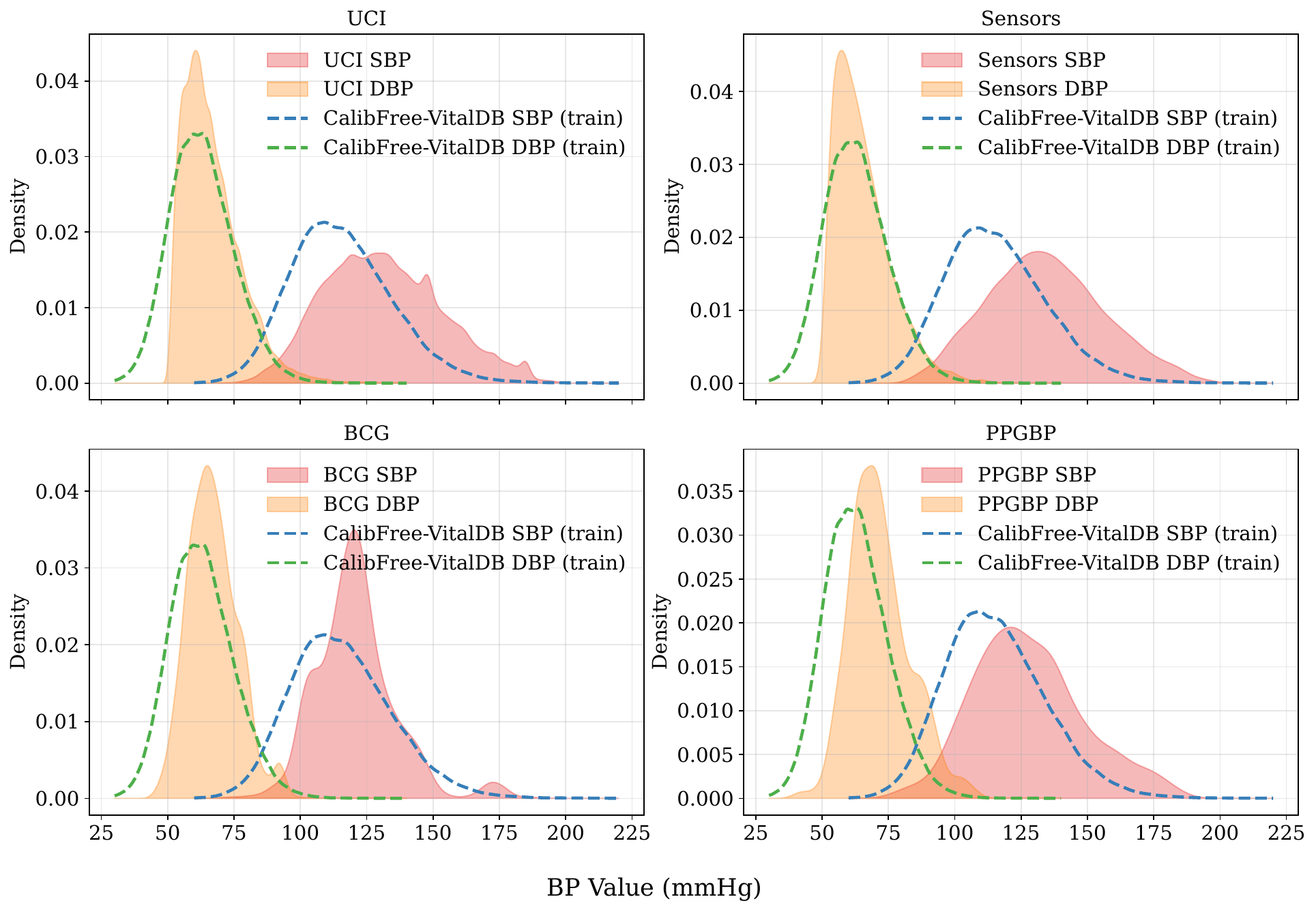}
    \caption{Comparison of SBP and DBP probability density distributions between ID training data
    (CalibFree-VitalDB) and OOD test datasets. Dashed curves indicate ID distributions, while filled curves indicate OOD distributions.}
    \label{fig:id_vs_ood_bp}
\end{figure}

Table~\ref{tab:emd_main} shows the Earth Mover’s Distance (EMD) \cite{pele2009fast} between the SBP and DBP distributions of the ID and OOD datasets. Sensors exhibits the largest distributional shift in SBP, followed by UCI, PPGBP, and BCG, indicating progressively decreasing deviation from the training distribution. Consistently, EMD values are lowest for BCG and highest for Sensors in SBP, while PPGBP shows the largest shift in DBP, with bootstrap-based confidence intervals highlighting substantially greater label-space variability under severe domain shift. Also, as shown based on the label distributions in Fig.~\ref{fig:id_vs_ood_bp}, the OOD datasets exhibit systematic distributional shifts relative to the CalibFree-VitalDB training data.
Note that this simple assessment is only supposed to convey a first assessment of the distribution shift based on the label distributions but does not cover distribution shift due to different input distributions.

\begin{table}[h]
\centering
\caption{EMD between ID and OOD datasets for SBP and DBP.}
\label{tab:emd_main}
\scalebox{1.10}{
\begin{tabular}{lcc}
\toprule
Dataset 
& SBP EMD [95\% CI] (mmHg) 
& DBP EMD [95\% CI] (mmHg) \\
\midrule
Sensors 
& 18.82 [18.39, 19.19] 
& 2.43 [2.22, 2.63] \\

UCI     
& 16.04 [15.91, 16.16] 
& 3.84 [3.77, 3.91] \\

PPGBP   
& 12.54 [10.97, 14.16] 
& 9.00 [8.08, 9.89] \\

BCG     
& 5.95 [5.62, 6.28] 
& 4.49 [4.22, 4.80] \\
\bottomrule
\end{tabular}
}
\end{table}

\subsection{ID predictive performance}
\label{sec:results_id_perf}

We first evaluated predictive performance under matched ID conditions on the CalibFree-Vital test set. Tables~\ref{tab:sbp_dropout_accuracy_winkler} and \ref{tab:dbp_dropout_accuracy_winkler} show that ID accuracy depends only moderately on the dropout rate. Overall, for MCD-based models, ID MAE varied only slightly across dropout rates (0\%, 4\%, and 40\%), with a maximum difference of 0.5~mmHg, observed for SBP in GNLL-based models. In more detail, for GNLL-based models, the lowest ID MAE was obtained at 40\% dropout for both SBP and DBP, reaching 12.33~mmHg and 7.90~mmHg, respectively, compared with 12.83~mmHg and 8.12~mmHg at 0\% dropout. A similar trend was observed for MSE-based models, where the best ID MAE was 12.40~mmHg for SBP and 7.90~mmHg for DBP at 40\% dropout.

DE further improved predictive performance through model averaging. As shown in Table~\ref{tab:de_merged_accuracy_uncertainty}, the MSE-based ensemble achieved the lowest ID MAE, with 12.18~mmHg for SBP and 7.73~mmHg for DBP, compared with 12.33~mmHg and 7.85~mmHg for the GNLL-based ensemble. In conclusion, under matched ID conditions, similar predictive accuracy was obtained for both GNLL- and MSE-based models, whereas ensemble averaging yielded the best performance in terms of point estimates.

\subsection{OOD predictive performance}
\label{sec:results_ood_perf}

We next evaluated predictive performance under OOD conditions on BCG, Sensors, UCI, and PPGBP. Aggregating results across random seeds using the median substantially mitigated the effects of underspecification in most cases (41 out of 48). Only in rare scenarios (specifically for the BCG dataset at dropout rates greater than 0\%) the OOD performance exceeds that observed under ID conditions. Overall, the strongest degradation was consistently observed on UCI and PPGBP. For instance, GNLL-based models reached an SBP MAE of 21.21~mmHg on UCI at 40\% dropout, while MSE-based models reached 22.44~mmHg on the same dataset. For DBP, a similar but less pronounced trend was observed, with BCG consistently remaining the least challenging OOD dataset.

DE improved robustness under shift, especially for SBP. Table~\ref{tab:de_merged_accuracy_uncertainty} shows that the GNLL-based deep ensemble reduced SBP MAE to 16.83~mmHg on PPGBP and remained competitive across all OOD datasets. For DBP, GNLL ensembles also achieved stable OOD performance, whereas MSE-based ensembles showed clearly larger degradation on Sensors and UCI. Overall, predictive performance deteriorated under domain shift, but DE helped mitigate this degradation, particularly for SBP.

\subsection{ID uncertainty reliability}
\label{sec:results_id_uq}

We next evaluated uncertainty reliability in the ID setting. Tables~\ref{tab:sbp_dropout_accuracy_winkler} and \ref{tab:dbp_dropout_accuracy_winkler} show that GNLL-based models already produced comparatively well-calibrated uncertainty estimates before recalibration. For SBP, Winkler1/Winkler2 improved slightly from 50.62/97.67 mmHg at 0\% dropout to 48.09/79.16 mmHg at 40\% dropout. For DBP, the corresponding values decreased from 31.53/52.69 mmHg to 30.31/48.41 mmHg, indicating overall stronger uncertainty reliability for DBP than for SBP. In contrast, MSE-based MCD yielded markedly poorer uncertainty quality in the ID setting. \deleted{because total uncertainty relied only on epistemic variability. As a result,} Winkler scores were substantially higher, for example, 74.42/466.06 mmHg for SBP at 4\% dropout and 46.97/287.82 mmHg for DBP at 4\% dropout.

Post-hoc recalibration improved ID uncertainty reliability for both model families, but the effect was much larger for MSE-based models. As shown in Tables~\ref{tab:sbp_recalibration_winkler} and \ref{tab:dbp_recalibration_winkler}, CP, TS, and IR only slightly refined the already reasonable benchmark scores for GNLL-based models, suggesting that recalibration mainly fine-tuned interval width rather than fundamentally changing uncertainty behaviour. In contrast, MSE-based models benefited strongly from recalibration. For example, for SBP at 40\% dropout, Winkler2 decreased from 402.63 mmHg in the benchmark to 82.04, 82.94, and 79.68 mmHg after CP, TS, and IR, respectively, with a similar pattern also observed for DBP. Comparing the recalibration methods, CP and TS were generally the most consistent, often yielding the lowest or near-lowest Winkler scores, whereas IR was also competitive and occasionally best, but somewhat more variable across settings. Overall, these results indicate that \added{parameter variance alone does not fully capture predictive uncertainty, especially for MSE-based models that do not explicitly learn aleatoric variance. Accordingly,} GNLL-based models already provide useful uncertainty estimates in the ID setting, while MSE-based models require recalibration to yield practically meaningful uncertainty intervals.

\subsection{OOD uncertainty reliability}
\label{sec:results_ood_uq}

Finally, we evaluated uncertainty reliability under domain shift. For GNLL-based models without dropout, uncertainty increased under OOD conditions, but often not enough to match the increase in prediction error. For example, in SBP, total uncertainty increased from 10.92~mmHg in ID to 12.58~mmHg on PPGBP, whereas MAE increased from 12.83 to 19.33~mmHg. This mismatch was reflected in worsening Winkler scores on the more challenging OOD datasets.

Increasing the dropout rate introduced epistemic uncertainty and improved uncertainty quality, although not always predictive accuracy. For GNLL-based SBP models on PPGBP, Winkler2 decreased from 192.67 mmHg at 0\% dropout to 166.47 mmHg at 40\% dropout, whereas on UCI, the corresponding MAE increased from 19.78 to 21.21~mmHg.

For MSE-based MCD models, unrecalibrated OOD uncertainty remained markedly less reliable, with very high Winkler2 values for SBP, for example, 824.60 mmHg on UCI at 4\% dropout. Post-hoc recalibration substantially improved these results and was often essential for making MSE-based uncertainty usable under shift. \added{This suggests that the uncertainty transformations learned during recalibration generalize to some extent beyond the calibration setting.} CP and TS generally showed the most consistent improvements, while IR was also competitive but somewhat more variable.

DE provided a complementary robustness mechanism. For GNLL-based DE, SBP epistemic uncertainty increased from 3.96 mmHg in ID to 10.69 mmHg on PPGBP while maintaining comparatively favourable Winkler2 values (130.67 mmHg). Recalibration further improved ensemble uncertainty quality, particularly for SBP and MSE-based ensembles, although some instability remained on PPGBP. Overall, these findings show that uncertainty reliability degrades under domain shift, but GNLL-based probabilistic modeling, ensembling, and recalibration each help improve uncertainty awareness in OOD settings.

\subsection{Statistical comparison of top-performing methods across datasets}

\begin{figure}[!ht]
    \centering
    \includegraphics[width=1.10\textwidth]{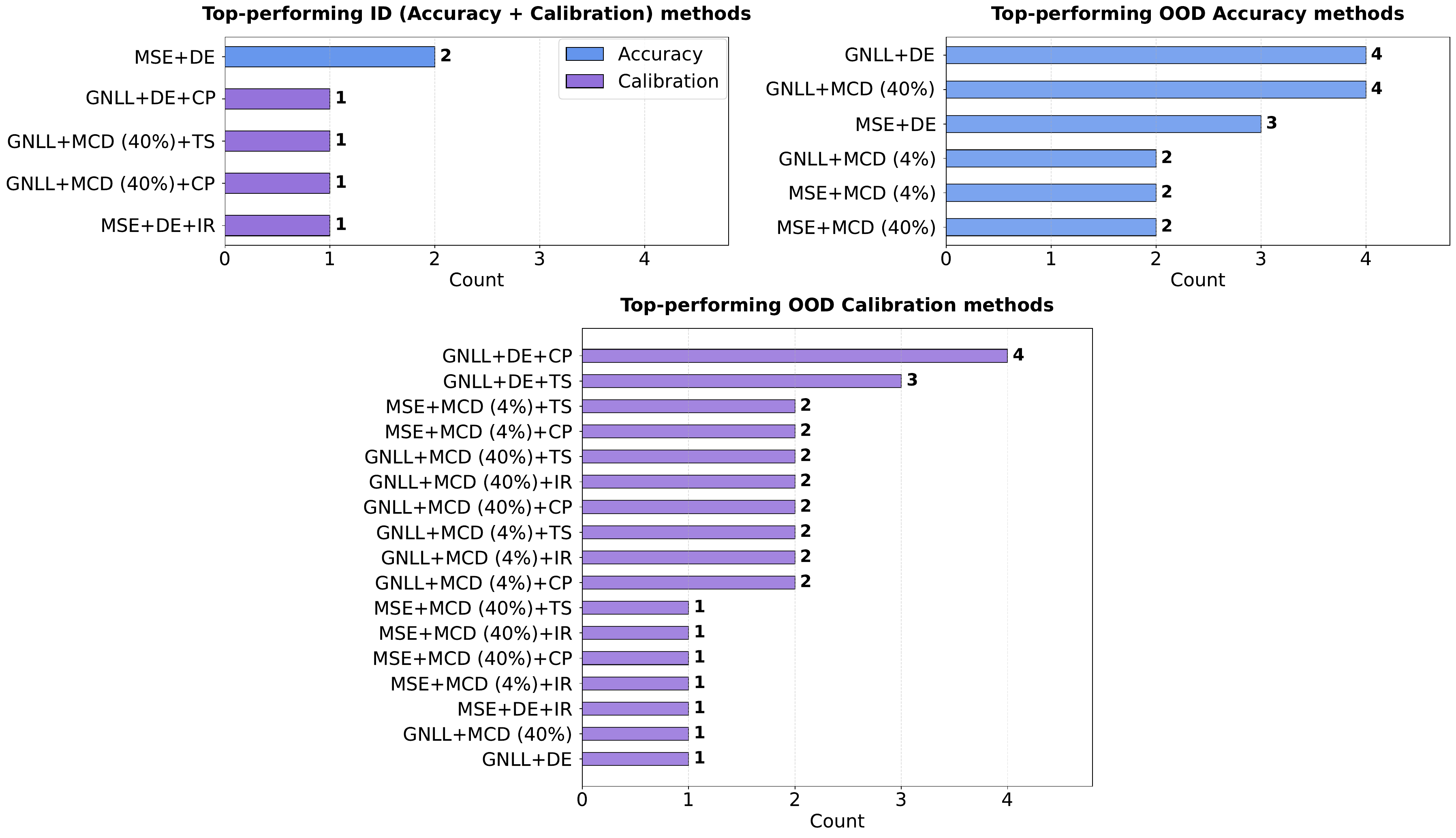}
    \caption{Bar-plot visualization of top-performing method frequencies across all slices. The left panel summarizes ID results by combining accuracy and calibration, with blue bars indicating accuracy and purple bars indicating calibration. The middle panel shows frequencies for OOD accuracy, and the right panel shows frequencies for OOD calibration. Bar length is proportional to how often a method appears across SBP and DBP analyses. Here, top-performing methods include the best-performing method and those whose 95\% paired bootstrap confidence interval for the difference to the best includes zero (Tier 1).}
    \label{fig:packed_circle_best_tier1}
\end{figure}

Tables~\ref{tab:sbp_best_tier1_tier2} and \ref{tab:dbp_best_tier1_tier2} summarize method performance across all slices using a tiered statistical comparison based on paired bootstrap analysis, while Fig.~\ref{fig:packed_circle_best_tier1} provides a global view of how frequently methods appear among the top-performing group, including both the best-performing methods and those that are not statistically distinguishable from the best (Tier~1). In the ID setting, MSE+DE achieved the best accuracy for both SBP and DBP, with MAEs of 12.18~mmHg and 7.73~mmHg, respectively. For calibration, GNLL+DE+CP achieved the best SBP result in ID (62.20~mmHg), whereas MSE+DE+IR achieved the best DBP result (39.06~mmHg).

For accuracy, ensemble-based approaches, particularly MSE+DE and GNLL+DE, frequently achieved the best or statistically comparable performance, especially in the ID and UCI settings. For example, MSE+DE was best for SBP in ID and UCI (12.18 and 18.79~mmHg), while GNLL+DE was best for DBP in Sensors and UCI (11.10 and 10.75~mmHg). MCD-based methods became more competitive in several OOD scenarios, including SBP on Sensors with GNLL+MCD (4\%) at 17.42~mmHg and DBP on PPGBP with GNLL+MCD (40\%) at 9.39~mmHg. For calibration, GNLL-based models combined with CP or TS frequently achieved the best performance or remained within Tier~1. GNLL+DE+CP achieved the best SBP calibration on ID, UCI, and PPGBP (62.20, 106.39, and 89.08~mmHg), whereas GNLL+MCD (4\%)+CP was best on BCG and Sensors (58.08 and 97.50~mmHg). For DBP, GNLL+DE+TS was best on Sensors and UCI (49.98 and 52.19~mmHg), while GNLL+DE was best on PPGBP (49.00~mmHg).


\clearpage

\newcolumntype{Y}{>{\raggedright\arraybackslash}X}
\begin{center}
\captionof{table}{Tiered comparison of methods for SBP across all datasets. For each slice and mode, \textbf{Tier~1} includes the best-performing model and all methods statistically compatible with the best, and \textbf{Tier~2} includes the three best-performing methods that are significantly worse than the best. Accuracy is reported as MAE (mmHg) and calibration as Winkler score (mmHg).}
\label{tab:sbp_best_tier1_tier2}

\footnotesize
\renewcommand{\arraystretch}{0.95}

\begin{tabularx}{\textwidth}{
l | l | l | Y | c | c | c
}
\toprule
\textbf{Slice} & \textbf{Mode} & \textbf{Tier} &
\makecell[l]{\textbf{Method}}
& \textbf{Metric (mmHg)} & $\boldsymbol{\Delta}$ vs \textbf{Best (mmHg)} & \textbf{95\% CI (mmHg)} \\
\midrule

\multirow{11}{*}{ID}
& \multirow{5}{*}{Accuracy}
& Tier 1 & \textbf{MSE+DE} & \textbf{12.18} & -- & -- \\
\cline{3-7}
&  & \multirow{3}{*}{Tier 2} & GNLL+DE & 12.33 & +0.15 & [+0.12, +0.17] \\
&  &  & MSE+MCD (40\%) & 12.38 & +0.20 & [+0.19, +0.22] \\
&  &  & GNLL+MCD (40\%) & 12.43 & +0.25 & [+0.23, +0.27] \\
\cline{2-7}
& \multirow{6}{*}{Calibration}
& \multirow{3}{*}{Tier 1} & \textbf{GNLL+DE+CP} & \textbf{62.20} & -- & -- \\
&  &  & GNLL+MCD (40\%)+TS & 62.21 & +0.01 & [-0.14, +0.16] \\
&  &  & GNLL+MCD (40\%)+CP & 62.26 & +0.06 & [-0.10, +0.22] \\
\cline{3-7}
&  & \multirow{3}{*}{Tier 2} & GNLL+DE+TS & 62.39 & +0.19 & [+0.13, +0.26] \\
&  &  & GNLL+MCD (40\%)+IR & 62.86 & +0.66 & [+0.49, +0.84] \\
&  &  & GNLL+DE+IR & 63.01 & +0.81 & [+0.70, +0.92] \\
\hline

\multirow{11}{*}{BCG}
& \multirow{5}{*}{Accuracy}
& Tier 1 & \textbf{MSE+MCD (40\%)} & \textbf{10.73} & -- & -- \\
\cline{3-7}
&  & \multirow{3}{*}{Tier 2} & MSE+MCD (4\%) & 11.05 & +0.32 & [+0.13, +0.52] \\
&  &  & GNLL+MCD (4\%) & 11.32 & +0.60 & [+0.38, +0.80] \\
&  &  & GNLL+MCD (40\%) & 11.84 & +1.11 & [+0.89, +1.37] \\
\cline{2-7}
& \multirow{6}{*}{Calibration}
& \multirow{3}{*}{Tier 1} & \textbf{GNLL+MCD (4\%)+CP} & \textbf{58.08} & -- & -- \\
&  &  & GNLL+MCD (4\%)+TS & 58.32 & +0.24 & [-0.14, +0.67] \\
&  &  & GNLL+MCD (4\%)+IR & 58.36 & +0.28 & [-0.20, +0.91] \\
\cline{3-7}
&  & \multirow{3}{*}{Tier 2} & GNLL+MCD (40\%)+CP & 61.05 & +2.97 & [+1.92, +4.12] \\
&  &  & MSE+MCD (4\%)+IR & 61.63 & +3.55 & [+1.81, +5.43] \\
&  &  & MSE+MCD (40\%)+IR & 61.91 & +3.83 & [+2.80, +4.99] \\
\hline

\multirow{11}{*}{Sensors}
& \multirow{6}{*}{Accuracy}
& \multirow{3}{*}{Tier 1}  & \textbf{GNLL+MCD (4\%)} & \textbf{17.42} & -- & -- \\
&  & & MSE+DE & 17.52 & +0.10 & [-0.56, +0.67] \\
&  &  & MSE+MCD (4\%) & 17.54 & +0.12 & [-0.08, +0.32] \\
\cline{3-7}
&  & \multirow{3}{*}{Tier 2} & GNLL+MCD (40\%) & 17.92 & +0.51 & [+0.28, +0.75] \\
&  &  & MSE+MCD (0\%) & 18.15 & +0.73 & [+0.14, +1.27] \\
&  &  & GNLL+DE & 18.32 & +0.90 & [+0.25, +1.51] \\
\cline{2-7}
& \multirow{6}{*}{Calibration}
& \multirow{3}{*}{Tier 1} & \textbf{GNLL+MCD (4\%)+CP} & \textbf{97.50} & -- & -- \\
&  &  & MSE+DE+IR & 99.94 & +2.44 & [-2.65, +8.31] \\
&  &  & GNLL+DE+CP & 100.82 & +3.32 & [-2.56, +8.49] \\
\cline{3-7}
&  & \multirow{3}{*}{Tier 2} & GNLL+MCD (4\%)+TS & 99.85 & +2.35 & [+2.00, +2.77] \\
&  &  & GNLL+MCD (4\%)+IR & 99.98 & +2.48 & [+1.69, +3.31] \\
&  &  & MSE+MCD (4\%)+CP & 101.04 & +3.54 & [+1.44, +5.59] \\
\hline

 \multirow{9}{*}{UCI}
& \multirow{5}{*}{Accuracy}
& Tier 1 & \textbf{MSE+DE} & \textbf{18.79} & -- & -- \\
\cline{3-7}
&  & \multirow{3}{*}{Tier 2} & GNLL+DE & 19.05 & +0.26 & [+0.24, +0.27] \\
&  &  & MSE+MCD (0\%) & 19.47 & +0.68 & [+0.68, +0.69] \\
&  &  & GNLL+MCD (0\%) & 20.51 & +1.71 & [+1.70, +1.73] \\
\cline{2-7}
& \multirow{5}{*}{Calibration}
& Tier 1 & \textbf{GNLL+DE+CP} & \textbf{106.39} & -- & -- \\
\cline{3-7}
&  & \multirow{3}{*}{Tier 2} & GNLL+DE+TS & 108.77 & +2.38 & [+2.34, +2.43] \\
&  &  & MSE+DE+IR & 110.48 & +4.09 & [+3.87, +4.31] \\
&  &  & GNLL+DE+IR & 112.71 & +6.32 & [+6.19, +6.48] \\
\hline

\multirow{20}{*}{PPGBP}
& \multirow{7}{*}{Accuracy}
& \multirow{4}{*}{Tier 1}  & \textbf{GNLL+DE} & \textbf{16.82} & -- & -- \\
&  & & MSE+DE & 17.13 & +0.31 & [-0.43, +1.02] \\
&  &  & GNLL+MCD (40\%) & 17.24 & +0.42 & [-2.07, +3.09] \\
&  &  & MSE+MCD (40\%) & 17.70 & +0.88 & [-1.53, +3.33] \\
\cline{3-7}
&  & \multirow{3}{*}{Tier 2} & GNLL+MCD (0\%) & 19.20 & +2.38 & [+2.10, +2.67] \\
&  &  & MSE+MCD (4\%) & 19.40 & +2.58 & [+0.53, +4.91] \\
&  &  & GNLL+MCD (4\%) & 21.93 & +5.11 & [+3.00, +7.55] \\
\cline{2-7}
& \multirow{12}{*}{Calibration}
& \multirow{9}{*}{Tier 1} & \textbf{GNLL+DE+CP} & \textbf{89.08} & -- & -- \\
&  &  & MSE+MCD (40\%)+CP & 95.44 & +6.36 & [-3.91, +19.11] \\
&  &  & MSE+MCD (40\%)+TS & 95.76 & +6.68 & [-4.51, +19.55] \\
&  &  & MSE+MCD (4\%)+CP & 101.20 & +12.12 & [-1.60, +27.03] \\
&  &  & MSE+MCD (4\%)+TS & 101.75 & +12.67 & [-1.29, +29.78] \\
&  &  & GNLL+MCD (40\%)+CP & 103.52 & +14.44 & [-6.78, +38.40] \\
&  &  & GNLL+MCD (40\%)+TS & 106.05 & +16.98 & [-3.19, +42.14] \\
&  &  & GNLL+MCD (40\%)+IR & 108.34 & +19.26 & [-2.89, +42.60] \\
&  &  & MSE+MCD (40\%)+IR & 108.70 & +19.62 & [-2.98, +46.48] \\
\cline{3-7}
&  & \multirow{3}{*}{Tier 2} & GNLL+DE+TS & 90.36 & +1.28 & [+0.29, +2.33] \\
&  &  & MSE+DE & 99.06 & +9.98 & [+5.31, +15.09] \\
&  &  & GNLL+DE & 99.56 & +10.48 & [+6.21, +14.82] \\
\bottomrule
\end{tabularx}
\end{center}

\clearpage

\section{Discussion}

\subsection{Distribution shift and OOD performance degradation}

\subsubsection{MCD-based models}

For MCD-based models, the point-prediction results in the present work were aggregated across five random seeds to reduce the effects of underspecification, thereby providing a more stable estimate of OOD behaviour. Under this more robust evaluation, the overall correlation between OOD SBP error and dataset dissimilarity quantified by EMD remained visible, consistent with the trend previously reported in \cite{moulaeifard2025mlppg}, although not in a strictly monotonic form. In particular, BCG consistently remained the least challenging OOD dataset, and UCI was typically among the most difficult, whereas the relative difficulty of Sensors and PPGBP varied across methods. 

\subsubsection{DE-based models}

For DE-based models, a similar overall tendency was observed: BCG generally remained the least challenging OOD dataset, whereas UCI and PPGBP were among the most difficult. At the same time, the ordering was again not perfectly monotonic with respect to EMD, indicating that label-space mismatch alone could not fully explain OOD difficulty. Compared with MCD-based models, DE-based models often showed slightly greater robustness under shift, but they were influenced by the same broader dataset-level differences.

\subsubsection{Overall interpretation}

Taken together, these findings suggest that EMD captures an important component of OOD shift, especially for SBP, but does not fully determine predictive degradation on its own. Rather, OOD difficulty appears to be shaped jointly by label-space differences and broader dataset-level factors, such as cohort composition, recording duration, and segment length. For instance, BCG was derived from only 40 subjects, whereas UCI included 10{,}793 subjects, and PPGBP additionally differed through its much shorter 2.1\,s segments (Table~\ref{tab:dataset}). This interpretation is consistent with prior work showing that strong ID performance alone is insufficient to establish deployment robustness, which must instead be evaluated explicitly on external datasets and under potential distribution shift \cite{han2025addressing,koch2024distribution}.

\subsection{Uncertainty modeling}

\subsubsection{MCD-based models}

For MCD-based models, one of the key findings of this study is that GNLL-based models consistently provided more reliable native uncertainty estimates than MSE-based models in both ID and OOD settings. This was particularly evident in the Winkler scores, where unrecalibrated MSE-based models showed markedly poorer uncertainty quality, whereas GNLL-based models remained substantially more reliable. This pattern is consistent with prior work distinguishing aleatoric and epistemic uncertainty \cite{kendall2017uncertainties,gal2016dropout}.

\subsubsection{DE-based models}

For DE-based models, native uncertainty estimates were generally stronger overall, particularly for GNLL-based ensembles, which combined robust predictive performance with comparatively reliable uncertainty quality. However, even in the DE setting, native probabilistic modeling did not fully prevent degradation under OOD shift. Uncertainty typically increased on external datasets, but not always enough to match the increase in prediction error, indicating that uncertainty quality still deteriorated under distribution shift.

\subsubsection{Overall interpretation}

Overall, these results show that native probabilistic modeling substantially improves uncertainty quality, but does not by itself guarantee reliable OOD uncertainty. Even for GNLL-based models, uncertainty estimates required explicit evaluation under external shift rather than being assumed to remain reliable by design. This interpretation is in line with the broader view that uncertainty estimates themselves must be externally validated, especially in high-stakes applications \cite{lambert2024trustworthy}.

\subsection{Effect of post-hoc recalibration}

\subsubsection{Recalibration for MCD-based models}

For MCD-based models, post-hoc recalibration substantially improved uncertainty quality, with the largest gains observed for MSE-based MCD. This is consistent with prior work showing that post-hoc calibration can meaningfully improve regression uncertainty estimates, especially when the underlying predictive uncertainty is initially poorly calibrated \cite{shafer2008tutorial,kuleshov2018accurate,levi2022evaluating}. Among the recalibration methods, CP and TS showed the most consistent improvements for MCD-based models, often yielding the lowest or near-lowest Winkler scores, whereas IR remained competitive but was somewhat more variable across datasets. The recalibration gains were also dataset-dependent: they were generally largest on the more difficult OOD datasets, especially UCI and Sensors, and smaller on BCG. This broadly aligns with the EMD analysis, suggesting that recalibration becomes more valuable as the target domain moves further away from the ID distribution, although the relationship was not strictly monotonic.

\subsubsection{Recalibration for DE-based models}

For DE-based models, recalibration also improved uncertainty quality, but the effect was typically smaller and less uniform than for MCD-based models. This is consistent with the stronger native uncertainty estimates already provided by ensemble-based predictive distributions \cite{lakshminarayanan2017simple,kuleshov2018accurate}. Within the DE setting, CP and TS again tended to be the most consistently effective strategies, while IR was occasionally competitive and in some cases among the best, but overall less stable across datasets. In our study, recalibration generally improved DE-based uncertainty on UCI and Sensors, while the gains were smaller on BCG. However, PPGBP exposed an important limitation: although GNLL-based DE remained competitive after recalibration, CP and TS could become unstable for MSE-based DE, substantially worsening Winkler scores in that setting.

\subsubsection{Overall interpretation}

Taken together, these results show that the effect of post-hoc recalibration \added{under ID and OOD conditions, as assessed using interval-based Winkler scores,} depends jointly on the base uncertainty model, the target dataset, and the recalibration strategy itself. Across settings, CP and TS showed the most consistent improvements overall, whereas IR remained competitive and was occasionally the best. This interpretation is consistent with the broader calibration literature: CP is attractive because it provides coverage guarantees under exchangeability, TS is often effective because of its simplicity and stability, and IR can be more flexible but also more sensitive to the calibration set \cite{guo2017calibration,shafer2008tutorial,kuleshov2018accurate}. Overall, our findings suggest that recalibration is most valuable when native uncertainty quality is weak, but its effectiveness under OOD shift also depends on \replaced{whether the relationship learned on the ID calibration set between predicted uncertainty and empirical prediction error remains valid in the external target dataset.}{how well the ID calibration set transfers to the external target domain.}

\subsection{Practical implications, limitations, and future directions}

Taken together, our results show that DE was generally strongest for predictive robustness, whereas GNLL-based models, particularly when combined with recalibration, were often strongest for uncertainty reliability. This is practically important for cuffless BP estimation, where real-world deployment requires not only strong average accuracy, but also robust and trustworthy uncertainty estimates across heterogeneous external settings \cite{yang2025cuffless}.

This study also has limitations. For instance, the OOD analysis was restricted to a selected set of public datasets, each with its own constraints in terms of cohort composition, recording duration, segment length, and acquisition conditions. Also, shift characterization in this work relied mainly on BP label distributions, which capture one important aspect of dataset mismatch but do not reflect other factors such as signal characteristics, acquisition conditions, or cohort differences. 

Future work should therefore extend external validation to a broader range of datasets and real-world measurement scenarios, characterize domain shift beyond BP marginals alone, and investigate hybrid or adaptive uncertainty strategies that are better suited to physiological time-series deployment under evolving conditions \cite{lambert2024trustworthy,yang2025cuffless}. In particular, leveraging large-scale and more diverse PPG benchmark datasets, such as the MIMIC-III-Ext-PPG dataset \cite{PhysioNet-mimic-iii-ext-ppg-1.1.0,moulaeifard2026mimic,moulaeifard2026deriving}, could enable more comprehensive evaluation of both ID and OOD generalization compared to the limited ID datasets used in the current study. \added{Finally, it is worth noting that the UQ evaluation in this work was based on Winkler score, i.e., a coverage-based metric. We acknowledge a large number of complementary, global but also local, UQ metrics as collected in \cite{bench2026systematic}, which could be also applied to this setting for more comprehensive insights.}

\section{Conclusion}

In this work, we systematically evaluated predictive performance and uncertainty reliability for PPG-based BP estimation under both ID and OOD conditions. While predictive accuracy remained relatively stable in the ID setting, performance and uncertainty reliability deteriorated under domain shift, highlighting that strong ID results alone are insufficient to assess real-world robustness. Overall, GNLL-based models provided more meaningful native uncertainty estimates than MSE-based models, whereas DE were often the strongest choice for predictive accuracy.

Post-hoc recalibration further improved uncertainty quality, with the largest benefits observed for MSE-based models. Taken together, these findings emphasize that robust cuffless BP estimation requires evaluating both point prediction and uncertainty behaviour across diverse external datasets, and that probabilistic modeling, ensembling, and recalibration each provide complementary benefits for improving reliability under distribution shift.

\section*{Data availability statement}
This work is exclusively based on publicly available datasets. The codes that support the findings of this study are openly available at the following URL/DOI: \url{https://github.com/AI4HealthUOL/UQ_OOD_PPG}

\section*{Acknowledgments}
The project (22HLT01 QUMPHY) has received funding from the European Partnership on Metrology, co-financed from the European Union’s Horizon Europe Research and Innovation Programme and by the Participating States. Funding for NPL was provided by Innovate UK under the Horizon Europe Guarantee Extension, grant number 10084125.

\bibliographystyle{unsrt}  
\bibliography{references}  

\clearpage
\section*{Appendix}

This appendix provides the detailed numerical results that complement the main findings presented in the main text. Specifically, Tables~\ref{tab:sbp_dropout_accuracy_winkler}--\ref{tab:dbp_recalibration_winkler} report the effects of loss function, dropout rate, and post-hoc recalibration on predictive performance and uncertainty reliability for SBP and DBP. Tables~\ref{tab:de_merged_accuracy_uncertainty} and \ref{tab:de_merged_recalibration} summarize the corresponding results for deep ensembles. Finally, Tables~\ref{tab:sbp_best_tier1_tier2} and \ref{tab:dbp_best_tier1_tier2} provide the tiered statistical comparison of methods across all evaluation slices.

\begin{table*}[t]
\centering
\captionsetup{skip=8pt}
\caption{Effect of loss function and dropout rate on SBP prediction performance and uncertainty reliability. Values are reported as median (IQR) across runs. Total uncertainty is computed as the square root of the sum of epistemic and aleatoric variances.}
\label{tab:sbp_dropout_accuracy_winkler}
\renewcommand{\arraystretch}{1.20}

\begin{tabular}{l c|ccc|cc}
\hline
\textbf{Test Dataset} & \textbf{MAE (mmHg)}~$\downarrow$ & \multicolumn{3}{c|}{\textbf{Uncertainty (mmHg)}~$\uparrow$} & \textbf{Winkler Score1} & \textbf{Winkler Score2} \\
\cline{3-5}
& & Epistemic & Aleatoric & Total & \textbf{(mmHg)}~$\downarrow$ & \textbf{(mmHg)}~$\downarrow$ \\
\hline
\multicolumn{7}{l}{\textbf{GNLL}} \\
\hline
\multicolumn{7}{l}{\textbf{\textit{Dropout Rate = 0\%}}} \\
CalibFree-Vital (ID) & 12.83 (0.43) & -- & 10.92 (4.80) & 10.92 (4.80) & 50.62 (8.26) & 97.67 (78.14) \\
BCG (OOD) & 13.59 (1.94) & -- & 11.13 (4.24) & 11.13 (4.24) & 59.72 (7.42) & 137.42 (28.31) \\
Sensors (OOD) & 18.70 (3.28) & -- & 11.67 (4.49) & 11.67 (4.49) & 88.41 (26.15) & 271.55 (124.15) \\
UCI (OOD) & 19.78 (2.93) & -- & 12.20 (4.95) & 12.20 (4.95) & 94.19 (12.32) & 254.45 (118.19) \\
PPGBP (OOD) & 19.33 (0.60) & -- & 12.58 (4.51) & 12.58 (4.51) & 82.44 (10.33) & 192.67 (108.49) \\
\hline
\multicolumn{7}{l}{\textbf{\textit{Dropout Rate = 4\%}}} \\
CalibFree-Vital (ID) & 12.62 (0.15) & 1.23 (0.16) & 12.17 (5.46) & 12.22 (5.42) & 49.59 (8.45) & 85.33 (84.90) \\
BCG (OOD) & 11.16 (1.04) & 1.53 (0.27) & 11.83 (5.53) & 11.93 (5.40) & 45.87 (3.06) & 90.39 (51.31) \\
Sensors (OOD) & 17.54 (0.66) & 1.55 (0.11) & 13.82 (5.79) & 13.91 (5.70) & 74.73 (12.02) & 173.59 (134.94) \\
UCI (OOD) & 20.91 (1.45) & 1.64 (0.10) & 14.42 (6.43) & 14.53 (6.34) & 100.27 (8.68) & 284.51 (151.51) \\
PPGBP (OOD) & 20.46 (0.46) & 2.14 (0.56) & 14.72 (6.98) & 14.88 (6.75) & 83.95 (12.49) & 180.79 (147.82) \\
\hline
\multicolumn{7}{l}{\textbf{\textit{Dropout Rate = 40\%}}} \\
CalibFree-Vital (ID) & 12.33 (0.17) & 1.94 (0.28) & 12.74 (1.83) & 12.86 (1.79) & 48.09 (0.77) & 79.16 (8.09) \\
BCG (OOD) & 12.15 (1.06) & 2.20 (0.23) & 12.65 (1.61) & 12.85 (1.59) & 49.05 (3.37) & 87.14 (9.82) \\
Sensors (OOD) & 17.88 (1.02) & 2.28 (0.70) & 13.25 (2.11) & 13.40 (2.27) & 75.50 (3.92) & 168.82 (28.18) \\
UCI (OOD) & 21.21 (0.31) & 2.75 (0.53) & 14.62 (2.04) & 15.05 (2.21) & 89.04 (4.11) & 207.51 (39.85) \\
PPGBP (OOD) & 17.09 (0.41) & 3.34 (0.53) & 14.43 (1.23) & 14.82 (1.29) & 74.42 (3.68) & 166.47 (22.45) \\
\hline
\multicolumn{7}{l}{\textbf{MSE}} \\
\hline
\multicolumn{7}{l}{\textbf{\textit{Dropout Rate = 0\%}}} \\
CalibFree-Vital (ID) & 12.46 (0.07) & -- & -- & -- & -- & -- \\
BCG (OOD) & 13.83 (0.91) & -- & -- & -- & -- & -- \\
Sensors (OOD) & 18.04 (1.03) & -- & -- & -- & -- & -- \\
UCI (OOD) & 19.41 (0.74) & -- & -- & -- & -- & -- \\
PPGBP (OOD) & 19.49 (5.74) & -- & -- & -- & -- & -- \\
\hline
\multicolumn{7}{l}{\textbf{\textit{Dropout Rate = 4\%}}} \\
CalibFree-Vital (ID) & 12.53 (0.08) & 1.05 (0.01) & -- & 1.05 (0.01) & 74.42 (0.55) & 466.06 (9.10) \\
BCG (OOD) & 10.87 (0.86) & 1.29 (0.13) & -- & 1.29 (0.13) & 63.25 (5.14) & 377.90 (31.25) \\
Sensors (OOD) & 17.02 (0.23) & 1.23 (0.04) & -- & 1.23 (0.04) & 102.26 (1.28) & 651.72 (7.04) \\
UCI (OOD) & 21.05 (1.09) & 1.26 (0.02) & -- & 1.26 (0.02) & 127.45 (6.74) & 824.60 (45.65) \\
PPGBP (OOD) & 20.65 (5.33) & 1.63 (0.03) & -- & 1.63 (0.03) & 123.43 (32.52) & 772.95 (219.90) \\
\hline
\multicolumn{7}{l}{\textbf{\textit{Dropout Rate = 40\%}}} \\
CalibFree-Vital (ID) & 12.40 (0.20) & 1.92 (0.10) & -- & 1.92 (0.10) & 70.39 (0.32) & 402.63 (7.10) \\
BCG (OOD) & 10.64 (0.95) & 2.50 (0.18) & -- & 2.50 (0.18) & 57.87 (3.65) & 305.71 (6.89) \\
Sensors (OOD) & 17.77 (2.10) & 2.53 (0.31) & -- & 2.53 (0.31) & 103.71 (12.63) & 623.67 (80.25) \\
UCI (OOD) & 22.44 (2.89) & 2.60 (0.27) & -- & 2.60 (0.27) & 132.50 (17.42) & 798.17 (129.86) \\
PPGBP (OOD) & 17.91 (1.34) & 3.04 (0.25) & -- & 3.04 (0.25) & 101.80 (5.77) & 541.34 (51.26) \\
\hline
\end{tabular}
\end{table*}

\begin{table*}[t]
\centering
\captionsetup{skip=8pt}
\caption{Effect of loss function and dropout rate on DBP prediction performance and uncertainty reliability. Values are reported as median (IQR) across runs. Total uncertainty is computed as the square root of the sum of epistemic and aleatoric variances.}
\label{tab:dbp_dropout_accuracy_winkler}
\renewcommand{\arraystretch}{1.12}

\begin{tabular}{l c|ccc|cc}
\hline
\textbf{Test Dataset} & \textbf{MAE (mmHg)}~$\downarrow$ & \multicolumn{3}{c|}{\textbf{Uncertainty (mmHg)}~$\uparrow$} & \textbf{Winkler Score1} & \textbf{Winkler Score2} \\
\cline{3-5}
& & Epistemic & Aleatoric & Total &  \textbf{(mmHg)}~$\downarrow$ &  \textbf{(mmHg)}~$\downarrow$\\
\hline
\multicolumn{7}{l}{\textbf{GNLL}} \\
\hline
\multicolumn{7}{l}{\textbf{\textit{Dropout Rate = 0\%}}} \\
CalibFree-Vital (ID) & 8.12 (0.24) & -- & 8.45 (1.19) & 8.45 (1.19) & 31.53 (4.21) & 52.69 (14.90) \\
BCG (OOD) & 8.78 (1.17) & -- & 8.83 (0.89) & 8.83 (0.89) & 34.89 (4.41) & 60.34 (11.23) \\
Sensors (OOD) & 12.30 (0.81) & -- & 9.19 (1.59) & 9.19 (1.59) & 47.00 (2.37) & 87.67 (19.22) \\
UCI (OOD) & 11.86 (0.58) & -- & 9.51 (1.57) & 9.51 (1.57) & 47.32 (2.04) & 87.65 (6.18) \\
PPGBP (OOD) & 10.92 (0.66) & -- & 10.22 (0.84) & 10.22 (0.84) & 43.43 (3.52) & 76.56 (13.90) \\
\hline
\multicolumn{7}{l}{\textbf{\textit{Dropout Rate = 4\%}}} \\
CalibFree-Vital (ID) & 8.08 (0.14) & 0.79 (0.16) & 9.12 (2.52) & 9.16 (2.50) & 30.90 (4.15) & 49.80 (25.88) \\
BCG (OOD) & 7.37 (0.37) & 0.93 (0.23) & 8.86 (2.36) & 8.91 (2.31) & 29.97 (2.23) & 53.03 (9.18) \\
Sensors (OOD) & 11.09 (3.57) & 0.93 (0.28) & 10.35 (2.91) & 10.40 (2.86) & 49.17 (24.08) & 111.46 (77.30) \\
UCI (OOD) & 12.14 (2.21) & 1.02 (0.25) & 10.85 (3.07) & 10.90 (3.02) & 55.97 (13.57) & 94.34 (80.33) \\
PPGBP (OOD) & 12.13 (5.45) & 1.29 (0.33) & 11.12 (2.90) & 11.20 (2.82) & 54.09 (22.54) & 103.15 (103.83) \\
\hline
\multicolumn{7}{l}{\textbf{\textit{Dropout Rate = 40\%}}} \\
CalibFree-Vital (ID) & 7.90 (0.14) & 1.14 (0.05) & 8.90 (0.62) & 8.98 (0.70) & 30.31 (0.35) & 48.41 (0.08) \\
BCG (OOD) & 7.65 (1.05) & 1.44 (0.28) & 8.96 (0.41) & 9.05 (0.36) & 30.29 (1.04) & 53.89 (6.56) \\
Sensors (OOD) & 11.45 (0.33) & 1.39 (0.54) & 9.41 (1.30) & 9.61 (1.29) & 45.64 (4.54) & 72.02 (31.31) \\
UCI (OOD) & 10.74 (0.47) & 1.89 (0.45) & 9.76 (0.91) & 9.98 (0.73) & 42.08 (3.08) & 75.40 (10.27) \\
PPGBP (OOD) & 9.09 (1.47) & 2.06 (0.41) & 10.05 (1.85) & 10.22 (1.99) & 35.82 (4.97) & 66.41 (10.96) \\
\hline
\multicolumn{7}{l}{\textbf{MSE}} \\
\hline
\multicolumn{7}{l}{\textbf{\textit{Dropout Rate = 0\%}}} \\
CalibFree-Vital (ID) & 7.90 (0.17) & -- & -- & -- & -- & -- \\
BCG (OOD) & 9.86 (1.46) & -- & -- & -- & -- & -- \\
Sensors (OOD) & 14.48 (1.21) & -- & -- & -- & -- & -- \\
UCI (OOD) & 12.88 (0.54) & -- & -- & -- & -- & -- \\
PPGBP (OOD) & 11.37 (8.68) & -- & -- & -- & -- & -- \\
\hline
\multicolumn{7}{l}{\textbf{\textit{Dropout Rate = 4\%}}} \\
CalibFree-Vital (ID) & 7.99 (0.12) & 0.78 (0.04) & -- & 0.78 (0.04) & 46.97 (0.69) & 287.82 (5.23) \\
BCG (OOD) & 9.03 (0.22) & 1.00 (0.05) & -- & 1.00 (0.05) & 53.01 (1.56) & 319.89 (15.75) \\
Sensors (OOD) & 12.57 (3.90) & 0.94 (0.04) & -- & 0.94 (0.04) & 75.20 (24.35) & 474.94 (166.33) \\
UCI (OOD) & 11.75 (3.50) & 0.96 (0.04) & -- & 0.96 (0.04) & 70.11 (21.72) & 440.42 (147.08) \\
PPGBP (OOD) & 10.08 (1.34) & 1.23 (0.09) & -- & 1.23 (0.09) & 58.22 (8.92) & 343.31 (67.02) \\
\hline
\multicolumn{7}{l}{\textbf{\textit{Dropout Rate = 40\%}}} \\
CalibFree-Vital (ID) & 7.90 (0.10) & 1.38 (0.18) & -- & 1.38 (0.18) & 44.30 (1.17) & 244.76 (14.70) \\
BCG (OOD) & 7.83 (1.36) & 1.77 (0.18) & -- & 1.77 (0.18) & 43.32 (8.42) & 232.37 (56.35) \\
Sensors (OOD) & 11.31 (2.05) & 1.84 (0.30) & -- & 1.84 (0.30) & 63.79 (10.91) & 354.83 (51.99) \\
UCI (OOD) & 10.72 (1.16) & 1.93 (0.29) & -- & 1.93 (0.29) & 59.98 (5.49) & 341.10 (29.65) \\
PPGBP (OOD) & 8.75 (1.58) & 2.32 (0.11) & -- & 2.32 (0.11) & 46.64 (9.63) & 231.94 (42.78) \\
\hline
\end{tabular}
\end{table*}

\begin{table*}[t]
\centering
\captionsetup{skip=8pt}
\caption{Effect of post-hoc recalibration on SBP uncertainty quality. Values are reported as median (IQR) across runs. Benchmark denotes the original unrecalibrated cases.}
\label{tab:sbp_recalibration_winkler}
\renewcommand{\arraystretch}{1.12}
\scalebox{0.76}{
\begin{tabular}{l|cc|cc|cc|cc}
\hline
\textbf{Test Dataset} & \multicolumn{2}{c|}{\textbf{Benchmark}} & \multicolumn{2}{c|}{\textbf{CP}} & \multicolumn{2}{c|}{\textbf{TS}} & \multicolumn{2}{c}{\textbf{IR}} \\
\cline{2-9}
 & Winkler Score1 & Winkler Score2 & Winkler Score1 & Winkler Score2 & Winkler Score1 & Winkler Score2 & Winkler Score1 & Winkler Score2 \\
 & (mmHg)~$\downarrow$ & (mmHg)~$\downarrow$ & (mmHg)~$\downarrow$ & (mmHg)~$\downarrow$ & (mmHg)~$\downarrow$ & (mmHg)~$\downarrow$ & (mmHg)~$\downarrow$ & (mmHg)~$\downarrow$ \\
\hline
\multicolumn{9}{l}{\textbf{GNLL}} \\
\hline
\multicolumn{9}{l}{\textbf{\textit{Dropout Rate = 0\%}}} \\
CalibFree-Vital (ID) & 50.62 (8.26) & 97.67 (78.14) & 49.84 (2.55) & 78.95 (2.58) & 49.27 (2.30) & 80.06 (2.72) & 49.33 (2.06) & 80.68 (4.39) \\
BCG (OOD) & 59.72 (7.42) & 137.42 (28.31) & 54.76 (9.00) & 88.62 (23.05) & 54.32 (8.94) & 91.30 (27.49) & 53.11 (8.57) & 84.90 (21.49) \\
Sensors (OOD) & 88.41 (26.15) & 271.55 (124.15) & 76.86 (14.80) & 138.74 (33.33) & 75.08 (13.52) & 147.11 (36.20) & 77.69 (14.40) & 155.29 (39.93) \\
UCI (OOD) & 94.19 (12.32) & 254.45 (118.19) & 82.07 (14.86) & 152.43 (36.53) & 80.17 (13.53) & 162.80 (40.70) & 83.08 (13.89) & 178.08 (44.94) \\
PPGBP (OOD) & 82.44 (10.33) & 192.67 (108.49) & 79.12 (3.12) & 150.14 (26.10) & 78.06 (3.26) & 156.23 (24.79) & 80.56 (11.31) & 173.97 (32.96) \\
\hline
\multicolumn{9}{l}{\textbf{\textit{Dropout Rate = 4\%}}} \\
CalibFree-Vital (ID) & 49.59 (8.45) & 85.33 (84.90) & 49.03 (0.69) & 77.38 (0.64) & 48.62 (0.70) & 78.57 (0.83) & 48.86 (0.68) & 80.06 (0.68) \\
BCG (OOD) & 45.87 (3.06) & 90.39 (51.31) & 43.02 (2.55) & 70.49 (1.27) & 43.37 (2.45) & 70.85 (1.51) & 43.91 (2.41) & 69.62 (6.94) \\
Sensors (OOD) & 74.73 (12.02) & 173.59 (134.94) & 70.96 (4.09) & 123.44 (6.37) & 69.71 (3.40) & 130.28 (6.37) & 71.04 (4.11) & 130.17 (8.00) \\
UCI (OOD) & 100.27 (8.68) & 284.51 (151.51) & 86.80 (5.82) & 166.05 (19.39) & 84.78 (6.23) & 174.85 (20.41) & 86.18 (2.97) & 177.57 (7.75) \\
PPGBP (OOD) & 83.95 (12.49) & 180.79 (147.82) & 80.48 (3.90) & 149.61 (13.87) & 79.38 (4.38) & 154.59 (20.47) & 79.76 (5.33) & 159.92 (14.65) \\
\hline
\multicolumn{9}{l}{\textbf{\textit{Dropout Rate = 40\%}}} \\
CalibFree-Vital (ID) & 48.09 (0.77) & 79.16 (8.09) & 47.65 (0.71) & 76.40 (1.79) & 47.33 (0.80) & 76.65 (1.09) & 47.61 (0.60) & 78.55 (1.86) \\
BCG (OOD) & 49.05 (3.37) & 87.14 (9.82) & 48.21 (3.05) & 72.05 (5.51) & 47.88 (3.07) & 75.07 (8.44) & 48.02 (3.34) & 76.49 (12.10) \\
Sensors (OOD) & 75.50 (3.92) & 168.82 (28.18) & 74.74 (4.01) & 143.13 (3.32) & 73.33 (3.68) & 151.11 (9.88) & 72.96 (4.09) & 151.54 (10.56) \\
UCI (OOD) & 89.04 (4.11) & 207.51 (39.85) & 89.05 (3.29) & 174.34 (6.50) & 87.27 (2.61) & 185.62 (13.54) & 87.16 (3.03) & 184.14 (23.52) \\
PPGBP (OOD) & 74.42 (3.68) & 166.47 (22.45) & 73.26 (1.18) & 139.16 (5.90) & 72.51 (1.39) & 146.06 (8.34) & 72.66 (0.61) & 149.10 (14.29) \\
\hline
\multicolumn{9}{l}{\textbf{MSE}} \\
\hline
\multicolumn{9}{l}{\textbf{\textit{Dropout Rate = 0\%}}} \\
CalibFree-Vital (ID) & -- & -- & -- & -- & -- & -- & -- & -- \\
BCG (OOD) & -- & -- & -- & -- & -- & -- & -- & -- \\
Sensors (OOD) & -- & -- & -- & -- & -- & -- & -- & -- \\
UCI (OOD) & -- & -- & -- & -- & -- & -- & -- & -- \\
PPGBP (OOD) & -- & -- & -- & -- & -- & -- & -- & -- \\
\hline
\multicolumn{9}{l}{\textbf{\textit{Dropout Rate = 4\%}}} \\
CalibFree-Vital (ID) & 74.42 (0.55) & 466.06 (9.10) & 49.27 (0.71) & 81.45 (1.22) & 48.78 (0.56) & 81.67 (1.19) & 48.36 (0.41) & 80.54 (2.99) \\
BCG (OOD) & 63.25 (5.14) & 377.90 (31.25) & 47.36 (3.74) & 80.72 (8.64) & 48.57 (4.16) & 79.63 (9.50) & 45.65 (3.72) & 72.48 (14.50) \\
Sensors (OOD) & 102.26 (1.28) & 651.72 (7.04) & 68.56 (2.06) & 123.94 (11.89) & 67.55 (2.14) & 127.19 (11.31) & 69.52 (1.58) & 149.81 (8.73) \\
UCI (OOD) & 127.45 (6.74) & 824.60 (45.65) & 88.22 (5.59) & 174.36 (17.70) & 86.42 (5.40) & 180.52 (16.57) & 90.41 (5.43) & 224.39 (16.15) \\
PPGBP (OOD) & 123.43 (32.52) & 772.95 (219.90) & 80.45 (19.25) & 126.35 (28.02) & 79.27 (16.64) & 130.12 (31.76) & 86.05 (22.79) & 165.93 (83.38) \\
\hline
\multicolumn{9}{l}{\textbf{\textit{Dropout Rate = 40\%}}} \\
CalibFree-Vital (ID) & 70.39 (0.32) & 402.63 (7.10) & 48.65 (0.33) & 82.04 (2.00) & 48.25 (0.43) & 82.94 (2.11) & 47.62 (0.54) & 79.68 (1.84) \\
BCG (OOD) & 57.87 (3.65) & 305.71 (6.89) & 45.74 (1.00) & 79.97 (3.92) & 47.57 (1.22) & 77.19 (4.60) & 44.72 (2.44) & 70.68 (18.74) \\
Sensors (OOD) & 103.71 (12.63) & 623.67 (80.25) & 72.88 (9.27) & 133.64 (14.90) & 71.89 (8.81) & 138.18 (18.33) & 73.49 (9.80) & 157.16 (31.31) \\
UCI (OOD) & 132.50 (17.42) & 798.17 (129.86) & 92.66 (14.37) & 166.35 (31.81) & 90.39 (14.29) & 175.98 (36.14) & 94.47 (14.82) & 215.94 (54.48) \\
PPGBP (OOD) & 101.80 (5.77) & 541.34 (51.26) & 69.59 (7.62) & 122.69 (10.39) & 69.94 (8.31) & 122.51 (10.72) & 71.38 (5.40) & 136.08 (25.98) \\
\hline
\end{tabular}
}
\end{table*}

\begin{table*}[t]
\centering
\captionsetup{skip=8pt}
\caption{Effect of post-hoc recalibration on DBP uncertainty quality. Values are reported as median (IQR) across runs. Benchmark denotes the original unrecalibrated cases.}
\label{tab:dbp_recalibration_winkler}
\renewcommand{\arraystretch}{1.12}
\scalebox{0.75}{
\begin{tabular}{l|cc|cc|cc|cc}
\hline
\textbf{Test Dataset} & \multicolumn{2}{c|}{\textbf{Benchmark}} & \multicolumn{2}{c|}{\textbf{CP}} & \multicolumn{2}{c|}{\textbf{TS}} & \multicolumn{2}{c}{\textbf{IR}} \\
\cline{2-9}
 & Winkler Score1 & Winkler Score2 & Winkler Score1 & Winkler Score2 & Winkler Score1 & Winkler2 & Winkler Score1 & Winkler Score2 \\
 & (mmHg)~$\downarrow$ & (mmHg)~$\downarrow$ & (mmHg)~$\downarrow$ & (mmHg)~$\downarrow$ & (mmHg)~$\downarrow$ & (mmHg)~$\downarrow$ & (mmHg)~$\downarrow$ & (mmHg)~$\downarrow$ \\
\hline
\multicolumn{9}{l}{\textbf{GNLL}} \\
\hline
\multicolumn{9}{l}{\textbf{\textit{Dropout Rate = 0\%}}} \\
CalibFree-Vital (ID) & 31.53 (4.21) & 52.69 (14.90) & 31.19 (1.89) & 50.33 (2.91) & 31.17 (2.51) & 49.95 (2.85) & 31.29 (1.79) & 50.43 (3.28) \\
BCG (OOD) & 34.89 (4.41) & 60.34 (11.23) & 33.27 (3.82) & 57.30 (4.21) & 33.27 (3.80) & 56.13 (6.42) & 33.53 (3.86) & 58.12 (4.31) \\
Sensors (OOD) & 47.00 (2.37) & 87.67 (19.22) & 48.09 (5.52) & 80.13 (13.99) & 48.36 (6.92) & 78.28 (10.86) & 48.11 (5.88) & 73.88 (7.81) \\
UCI (OOD) & 47.32 (2.04) & 87.65 (6.18) & 45.76 (2.08) & 81.68 (5.74) & 46.26 (3.80) & 79.26 (6.70) & 45.23 (2.33) & 73.88 (14.78) \\
PPGBP (OOD) & 43.43 (3.52) & 76.56 (13.90) & 42.08 (3.86) & 66.20 (13.21) & 42.08 (3.67) & 66.24 (12.45) & 44.63 (4.48) & 84.97 (17.74) \\
\hline
\multicolumn{9}{l}{\textbf{\textit{Dropout Rate = 4\%}}} \\
CalibFree-Vital (ID) & 30.90 (4.15) & 49.80 (25.88) & 30.90 (0.98) & 50.86 (0.92) & 30.91 (0.98) & 49.90 (0.69) & 31.03 (1.02) & 50.68 (0.86) \\
BCG (OOD) & 29.97 (2.23) & 53.03 (9.18) & 28.33 (2.76) & 44.78 (0.74) & 28.38 (2.82) & 44.31 (0.39) & 27.90 (2.98) & 42.67 (5.23) \\
Sensors (OOD) & 49.17 (24.08) & 111.46 (77.30) & 42.66 (16.31) & 66.75 (38.32) & 42.40 (16.24) & 64.93 (33.98) & 42.15 (13.73) & 65.90 (19.02) \\
UCI (OOD) & 55.97 (13.57) & 94.34 (80.33) & 49.72 (10.21) & 100.21 (23.30) & 49.48 (10.18) & 95.12 (18.36) & 50.05 (7.79) & 86.01 (10.24) \\
PPGBP (OOD) & 54.09 (22.54) & 103.15 (103.83) & 47.81 (22.14) & 86.71 (41.42) & 47.52 (21.81) & 83.55 (37.95) & 48.12 (20.47) & 87.49 (37.63) \\
\hline
\multicolumn{9}{l}{\textbf{\textit{Dropout Rate = 40\%}}} \\
CalibFree-Vital (ID) & 30.31 (0.35) & 48.41 (0.08) & 30.41 (0.20) & 48.43 (1.01) & 30.36 (0.22) & 47.95 (1.00) & 30.34 (0.29) & 48.73 (1.11) \\
BCG (OOD) & 30.29 (1.04) & 53.89 (6.56) & 30.49 (0.97) & 50.60 (2.73) & 30.56 (0.98) & 49.13 (3.36) & 30.37 (1.42) & 48.93 (2.14) \\
Sensors (OOD) & 45.64 (4.54) & 72.02 (31.31) & 43.90 (1.38) & 64.42 (5.01) & 43.43 (1.28) & 62.76 (3.89) & 43.55 (1.43) & 63.10 (12.30) \\
UCI (OOD) & 42.08 (3.08) & 75.40 (10.27) & 41.41 (2.35) & 70.13 (6.22) & 41.30 (2.42) & 69.95 (5.90) & 41.38 (2.60) & 71.42 (6.84) \\
PPGBP (OOD) & 35.82 (4.97) & 66.41 (10.96) & 35.52 (4.98) & 62.92 (11.60) & 35.47 (4.86) & 62.77 (10.72) & 36.72 (4.29) & 64.05 (6.23) \\
\hline
\multicolumn{9}{l}{\textbf{MSE}} \\
\hline
\multicolumn{9}{l}{\textbf{\textit{Dropout Rate = 0\%}}} \\
CalibFree-Vital (ID) & -- & -- & -- & -- & -- & -- & -- & -- \\
BCG (OOD) & -- & -- & -- & -- & -- & -- & -- & -- \\
Sensors (OOD) & -- & -- & -- & -- & -- & -- & -- & -- \\
UCI (OOD) & -- & -- & -- & -- & -- & -- & -- & -- \\
PPGBP (OOD) & -- & -- & -- & -- & -- & -- & -- & -- \\
\hline
\multicolumn{9}{l}{\textbf{\textit{Dropout Rate = 4\%}}} \\
CalibFree-Vital (ID) & 46.97 (0.69) & 287.82 (5.23) & 31.30 (0.55) & 50.74 (0.74) & 31.22 (0.65) & 50.42 (0.36) & 30.85 (0.64) & 49.14 (0.29) \\
BCG (OOD) & 53.01 (1.56) & 319.89 (15.75) & 37.78 (2.33) & 60.75 (1.90) & 38.02 (2.55) & 60.29 (2.25) & 37.22 (1.62) & 68.87 (5.55) \\
Sensors (OOD) & 75.20 (24.35) & 474.94 (166.33) & 48.16 (18.45) & 75.02 (44.46) & 47.40 (18.17) & 74.14 (41.55) & 49.92 (19.29) & 92.13 (61.03) \\
UCI (OOD) & 70.11 (21.72) & 440.42 (147.08) & 46.77 (13.85) & 85.18 (21.28) & 46.29 (13.48) & 84.38 (17.49) & 47.83 (16.09) & 96.76 (37.09) \\
PPGBP (OOD) & 58.22 (8.92) & 343.31 (67.02) & 39.27 (2.60) & 62.13 (1.15) & 39.40 (2.47) & 62.50 (2.25) & 39.77 (3.30) & 71.91 (20.55) \\
\hline
\multicolumn{9}{l}{\textbf{\textit{Dropout Rate = 40\%}}} \\
CalibFree-Vital (ID) & 44.30 (1.17) & 244.76 (14.70) & 30.89 (0.08) & 51.14 (2.56) & 30.85 (0.08) & 51.26 (2.34) & 30.69 (0.35) & 49.20 (1.71) \\
BCG (OOD) & 43.32 (8.42) & 232.37 (56.35) & 32.39 (1.48) & 49.76 (3.83) & 32.80 (1.28) & 50.52 (3.42) & 31.44 (3.48) & 53.57 (8.26) \\
Sensors (OOD) & 63.79 (10.91) & 354.83 (51.99) & 41.21 (6.25) & 59.92 (3.24) & 40.94 (6.01) & 59.80 (3.22) & 41.97 (9.61) & 61.64 (19.64) \\
UCI (OOD) & 59.98 (5.49) & 341.10 (29.65) & 41.10 (3.60) & 70.50 (3.19) & 41.02 (3.49) & 70.28 (3.07) & 41.73 (5.15) & 76.89 (10.83) \\
PPGBP (OOD) & 46.64 (9.63) & 231.94 (42.78) & 38.01 (1.96) & 66.65 (9.58) & 38.71 (2.23) & 67.44 (10.05) & 34.94 (1.66) & 55.31 (9.55) \\
\hline
\end{tabular}
}
\end{table*}

\begin{table*}[t]
\centering
\captionsetup{skip=8pt}
\caption{DE results for SBP and DBP prediction performance and uncertainty reliability. Total uncertainty is computed as the square root of the sum of epistemic and aleatoric variances.}
\label{tab:de_merged_accuracy_uncertainty}
\renewcommand{\arraystretch}{1.12}

\begin{tabular}{l c|ccc|cc}
\hline
\textbf{Test Dataset} & \textbf{MAE (mmHg)}~$\downarrow$ & \multicolumn{3}{c|}{\textbf{Uncertainty (mmHg)}~$\uparrow$} & \textbf{Winkler Score1} & \textbf{Winkler Score2} \\
\cline{3-5}
& & Epistemic & Aleatoric & Total & \textbf{(mmHg)}~$\downarrow$ & \textbf{(mmHg)}~$\downarrow$ \\
\hline
\multicolumn{7}{l}{\textbf{SBP}} \\
\hline
\multicolumn{7}{l}{\textbf{\textit{GNLL}}} \\
CalibFree-Vital (ID) & 12.33 & 3.96 & 10.75 & 11.59 & 49.06 & 88.57 \\
BCG (OOD) & 12.76 & 6.21 & 10.78 & 12.62 & 49.66 & 91.83 \\
Sensors (OOD) & 18.32 & 8.07 & 11.84 & 14.60 & 75.68 & 164.86 \\
UCI (OOD) & 19.05 & 8.87 & 12.40 & 15.59 & 78.69 & 172.67 \\
PPGBP (OOD) & 16.83 & 10.69 & 12.63 & 16.90 & 68.52 & 130.67 \\
\hline
\multicolumn{7}{l}{\textbf{\textit{MSE}}} \\
CalibFree-Vital (ID) & 12.18 & 3.02 & 0.00 & 3.02 & 65.48 & 329.28 \\
BCG (OOD) & 13.36 & 5.21 & 0.00 & 5.21 & 66.89 & 278.67 \\
Sensors (OOD) & 17.52 & 5.57 & 0.00 & 5.57 & 90.45 & 411.55 \\
UCI (OOD) & 18.79 & 5.94 & 0.00 & 5.94 & 97.25 & 444.44 \\
PPGBP (OOD) & 17.13 & 25.20 & 0.00 & 25.20 & 71.88 & 126.26 \\
\hline
\multicolumn{7}{l}{\textbf{DBP}} \\
\hline
\multicolumn{7}{l}{\textbf{\textit{GNLL}}} \\
CalibFree-Vital (ID) & 7.85 & 2.87 & 8.27 & 8.83 & 30.35 & 48.83 \\
BCG (OOD) & 7.91 & 4.38 & 8.44 & 9.64 & 31.38 & 52.58 \\
Sensors (OOD) & 11.10 & 5.38 & 9.30 & 10.88 & 41.36 & 59.94 \\
UCI (OOD) & 10.75 & 5.88 & 9.55 & 11.39 & 40.18 & 64.83 \\
PPGBP (OOD) & 9.45 & 7.07 & 9.80 & 12.32 & 36.92 & 61.31 \\
\hline
\multicolumn{7}{l}{\textbf{\textit{MSE}}} \\
CalibFree-Vital (ID) & 7.73 & 2.76 & 0.00 & 2.76 & 39.12 & 168.66 \\
BCG (OOD) & 9.84 & 5.27 & 0.00 & 5.27 & 44.36 & 140.09 \\
Sensors (OOD) & 15.04 & 5.44 & 0.00 & 5.44 & 74.25 & 287.47 \\
UCI (OOD) & 13.26 & 5.47 & 0.00 & 5.47 & 63.95 & 232.62 \\
PPGBP (OOD) & 12.85 & 12.11 & 0.00 & 12.11 & 49.49 & 87.01 \\
\hline
\end{tabular}
\end{table*}

\begin{table*}[t]
\centering
\captionsetup{skip=8pt}
\caption{Effect of post-hoc recalibration on DE uncertainty quality for SBP and DBP. Benchmark denotes the original unrecalibrated cases.}
\label{tab:de_merged_recalibration}
\renewcommand{\arraystretch}{1.12}
\scalebox{0.76}{
\begin{tabular}{l|cc|cc|cc|cc}
\hline
\textbf{Test Dataset} & \multicolumn{2}{c|}{\textbf{Benchmark}} & \multicolumn{2}{c|}{\textbf{CP}} & \multicolumn{2}{c|}{\textbf{TS}} & \multicolumn{2}{c}{\textbf{IR}} \\
\cline{2-9}
 & Winkler Score1 & Winkler Score2 & Winkler Score1 & Winkler Score2 & Winkler Score1 & Winkler Score2 & Winkler Score1 & Winkler Score2 \\
 & (mmHg)~$\downarrow$ & (mmHg)~$\downarrow$ & (mmHg)~$\downarrow$ & (mmHg)~$\downarrow$ & (mmHg)~$\downarrow$ & (mmHg)~$\downarrow$ & (mmHg)~$\downarrow$ & (mmHg)~$\downarrow$ \\
\hline
\multicolumn{9}{l}{\textbf{SBP}} \\
\hline
\multicolumn{9}{l}{\textbf{\textit{GNLL}}} \\
CalibFree-Vital (ID) & 49.06 & 88.57 & 48.18 & 76.23 & 47.68 & 77.12 & 47.91 & 78.13 \\
BCG (OOD) & 49.66 & 91.83 & 49.18 & 77.89 & 49.16 & 78.92 & 49.17 & 81.14 \\
Sensors (OOD) & 75.68 & 164.86 & 73.86 & 127.80 & 72.54 & 133.97 & 73.39 & 134.94 \\
UCI (OOD) & 78.69 & 172.67 & 76.99 & 135.78 & 75.79 & 141.75 & 77.88 & 147.55 \\
PPGBP (OOD) & 68.52 & 130.67 & 67.78 & 110.39 & 67.46 & 113.25 & 72.50 & 128.46 \\
\hline
\multicolumn{9}{l}{\textbf{\textit{MSE}}} \\
CalibFree-Vital (ID) & 65.48 & 329.28 & 49.93 & 96.03 & 50.74 & 96.02 & 47.33 & 80.46 \\
BCG (OOD) & 66.89 & 278.67 & 59.98 & 122.75 & 66.06 & 122.45 & 53.99 & 89.80 \\
Sensors (OOD) & 90.45 & 411.55 & 73.38 & 143.16 & 78.09 & 142.93 & 69.99 & 129.91 \\
UCI (OOD) & 97.25 & 444.44 & 79.15 & 155.86 & 84.17 & 155.64 & 75.50 & 145.44 \\
PPGBP (OOD) & 71.88 & 126.26 & 239.30 & 576.15 & 287.21 & 574.41 & 90.07 & 167.88 \\
\hline
\multicolumn{9}{l}{\textbf{DBP}} \\
\hline
\multicolumn{9}{l}{\textbf{\textit{GNLL}}} \\
CalibFree-Vital (ID) & 30.35 & 48.83 & 30.31 & 48.94 & 30.28 & 48.21 & 30.55 & 49.89 \\
BCG (OOD) & 31.38 & 52.58 & 31.36 & 52.65 & 31.35 & 52.18 & 32.06 & 54.16 \\
Sensors (OOD) & 41.36 & 59.94 & 41.17 & 60.11 & 41.04 & 58.86 & 42.47 & 67.06 \\
UCI (OOD) & 40.18 & 64.83 & 40.06 & 64.91 & 39.98 & 64.39 & 44.31 & 79.03 \\
PPGBP (OOD) & 36.92 & 61.31 & 36.96 & 61.30 & 36.99 & 61.51 & 47.89 & 89.22 \\
\hline
\multicolumn{9}{l}{\textbf{\textit{MSE}}} \\
CalibFree-Vital (ID) & 39.12 & 168.66 & 31.70 & 60.96 & 32.75 & 61.12 & 29.87 & 48.23 \\
BCG (OOD) & 44.36 & 140.09 & 42.67 & 88.61 & 48.17 & 90.38 & 38.02 & 68.42 \\
Sensors (OOD) & 74.25 & 287.47 & 52.00 & 92.59 & 54.00 & 94.27 & 59.30 & 103.43 \\
UCI (OOD) & 63.95 & 232.62 & 48.61 & 94.17 & 52.28 & 95.86 & 51.44 & 84.62 \\
PPGBP (OOD) & 49.49 & 87.01 & 85.20 & 202.98 & 103.74 & 207.15 & 50.99 & 86.40 \\
\hline
\end{tabular}
}
\end{table*}

\newcolumntype{Y}{>{\raggedright\arraybackslash}X}

\begin{table*}[ht]
\centering
\caption{Tiered comparison of methods for DBP across all datasets. For each slice and mode, \textbf{Tier~1} includes the best-performing model and all methods statistically compatible with the best, and \textbf{Tier~2} includes the three best-performing methods that are significantly worse than the best. Accuracy is reported as MAE (mmHg) and calibration as Winkler score (mmHg).}
\label{tab:dbp_best_tier1_tier2}
\footnotesize
\renewcommand{\arraystretch}{1.05}

\begin{tabularx}{\textwidth}{
l | l | l | Y | c | c | c
}
\toprule
\textbf{Slice} & \textbf{Mode} & \textbf{Tier} &
\makecell[l]{\textbf{Method}}
& \textbf{Metric (mmHg)} & $\boldsymbol{\Delta}$ vs \textbf{Best (mmHg)} & \textbf{95\% CI (mmHg)} \\
\midrule

\multirow{10}{*}{ID}
& \multirow{5}{*}{Accuracy}
& Tier 1 & \textbf{MSE+DE} & \textbf{7.73} & -- & -- \\
\cline{3-7}
&  & \multirow{3}{*}{Tier 2} & GNLL+DE & 7.85 & +0.12 & [+0.11, +0.14] \\
&  &  & GNLL+MCD (40\%) & 7.91 & +0.18 & [+0.17, +0.19] \\
&  &  & MSE+MCD (40\%) & 7.92 & +0.19 & [+0.18, +0.20] \\
\cline{2-7}
& \multirow{5}{*}{Calibration}
& Tier 1 & \textbf{MSE+DE+IR} & \textbf{39.06} & -- & -- \\
\cline{3-7}
&  & \multirow{3}{*}{Tier 2} & GNLL+DE+TS & 39.25 & +0.20 & [+0.07, +0.32] \\
&  &  & GNLL+MCD (40\%)+TS & 39.58 & +0.53 & [+0.41, +0.64] \\
&  &  & GNLL+DE & 39.60 & +0.54 & [+0.42, +0.67] \\
\hline

\multirow{10}{*}{BCG}
& \multirow{5}{*}{Accuracy}
& \multirow{2}{*}{Tier 1} & \textbf{GNLL+MCD (4\%)} & \textbf{7.37} & -- & -- \\
&  & & GNLL+MCD (40\%) & 7.51 & +0.14 & [-0.03, +0.29] \\
\cline{3-7}
&  & \multirow{3}{*}{Tier 2} & GNLL+DE & 7.91 & +0.54 & [+0.08, +1.07] \\
&  &  & MSE+MCD (40\%) & 8.45 & +1.08 & [+0.92, +1.24] \\
&  &  & GNLL+MCD (0\%) & 8.86 & +1.48 & [+1.03, +1.99] \\
\cline{2-7}
& \multirow{5}{*}{Calibration}
& \multirow{2}{*}{Tier 1} & \textbf{GNLL+MCD (4\%)+IR} & \textbf{36.99} & -- & -- \\
&  &  & GNLL+MCD (4\%)+TS & 37.12 & +0.13 & [-0.07, +0.30] \\
\cline{3-7}
&  & \multirow{3}{*}{Tier 2} & GNLL+MCD (4\%)+CP & 37.38 & +0.39 & [+0.20, +0.58] \\
&  &  & GNLL+MCD (40\%)+IR & 40.47 & +3.48 & [+2.01, +4.87] \\
&  &  & GNLL+MCD (40\%)+TS & 40.70 & +3.71 & [+2.36, +5.18] \\
\hline

\multirow{10}{*}{Sensors}
& \multirow{5}{*}{Accuracy}
& Tier 1 & \textbf{GNLL+DE} & \textbf{11.10} & -- & -- \\
\cline{3-7}
&  & \multirow{3}{*}{Tier 2} & GNLL+MCD (40\%) & 11.75 & +0.65 & [+0.31, +0.96] \\
&  &  & GNLL+MCD (0\%) & 11.95 & +0.85 & [+0.82, +0.88] \\
&  &  & MSE+MCD (40\%) & 12.32 & +1.22 & [+0.90, +1.51] \\
\cline{2-7}
& \multirow{5}{*}{Calibration}
& Tier 1 & \textbf{GNLL+DE+TS} & \textbf{49.98} & -- & -- \\
\cline{3-7}
&  & \multirow{3}{*}{Tier 2} & GNLL+DE+CP & 50.68 & +0.70 & [+0.60, +0.78] \\
&  &  & GNLL+DE & 50.68 & +0.70 & [+0.62, +0.78] \\
&  &  & GNLL+DE+IR & 54.79 & +4.81 & [+4.32, +5.29] \\
\hline

\multirow{11}{*}{UCI}
& \multirow{6}{*}{Accuracy}
& \multirow{2}{*}{Tier 1} & \textbf{GNLL+DE} & \textbf{10.75} & -- & -- \\
&  &  & GNLL+MCD (40\%) & 10.79 & +0.04 & [-0.00, +0.08] \\
\cline{3-7}
&  & \multirow{3}{*}{Tier 2} & MSE+MCD (40\%) & 11.22 & +0.47 & [+0.43, +0.51] \\
&  &  & GNLL+MCD (0\%) & 11.80 & +1.05 & [+1.04, +1.06] \\
&  &  & GNLL+MCD (4\%) & 12.22 & +1.47 & [+1.42, +1.51] \\
\cline{2-7}
& \multirow{5}{*}{Calibration}
& Tier 1 & \textbf{GNLL+DE+TS} & \textbf{52.19} & -- & -- \\
\cline{3-7}
&  & \multirow{3}{*}{Tier 2} & GNLL+DE+CP & 52.49 & +0.30 & [+0.28, +0.31] \\
&  &  & GNLL+DE & 52.51 & +0.32 & [+0.30, +0.33] \\
&  &  & GNLL+MCD (40\%)+TS & 57.18 & +4.99 & [+4.56, +5.42] \\
\hline

\multirow{21}{*}{PPGBP}
& \multirow{6}{*}{Accuracy}
& \multirow{3}{*}{Tier 1} & \textbf{GNLL+MCD (40\%)} & \textbf{9.39} & -- & -- \\
&  &  & GNLL+DE & 9.44 & +0.05 & [-1.23, +1.20] \\
&  &  & MSE+MCD (4\%) & 10.01 & +0.62 & [-0.06, +1.29] \\
\cline{3-7}
&  & \multirow{3}{*}{Tier 2} & GNLL+MCD (0\%) & 11.18 & +1.80 & [+0.61, +3.05] \\
&  &  & MSE+MCD (40\%) & 11.26 & +1.87 & [+1.36, +2.30] \\
&  &  & GNLL+MCD (4\%) & 11.81 & +2.42 & [+1.83, +2.98] \\
\cline{2-7}
& \multirow{13}{*}{Calibration}
& \multirow{10}{*}{Tier 1} & \textbf{GNLL+DE} & \textbf{49.00} & -- & -- \\
&  &  & GNLL+DE+CP & 49.01 & +0.01 & [-0.03, +0.05] \\
&  &  & GNLL+DE+TS & 49.13 & +0.13 & [-0.18, +0.39] \\
&  &  & GNLL+MCD (40\%)+TS & 49.52 & +0.52 & [-8.04, +10.94] \\
&  &  & GNLL+MCD (40\%)+CP & 49.91 & +0.91 & [-8.18, +12.05] \\
&  &  & MSE+MCD (4\%)+CP & 50.46 & +1.47 & [-3.30, +5.83] \\
&  &  & MSE+MCD (4\%)+TS & 50.89 & +1.89 & [-3.04, +6.07] \\
&  &  & GNLL+MCD (40\%)+IR & 51.49 & +2.49 & [-6.83, +12.96] \\
&  &  & GNLL+MCD (40\%) & 52.42 & +3.43 & [-5.79, +15.09] \\
&  &  & MSE+MCD (4\%)+IR & 55.15 & +6.15 & [-1.89, +15.60] \\
\cline{3-7}
&  & \multirow{3}{*}{Tier 2} & MSE+MCD (40\%)+CP & 63.05 & +14.05 & [+8.10, +21.19] \\
&  &  & GNLL+MCD (4\%)+TS & 63.21 & +14.21 & [+4.32, +24.71] \\
&  &  & MSE+MCD (40\%)+TS & 63.77 & +14.77 & [+8.95, +20.45] \\
\bottomrule
\end{tabularx}
\end{table*}

\end{document}